\documentclass[10pt,twocolumn,letterpaper]{article}

\usepackage{iccv}
\usepackage{times}
\usepackage{epsfig}
\usepackage{graphicx}
\usepackage{amsmath}
\usepackage{amssymb}
\usepackage{booktabs}
\usepackage[dvipsnames]{xcolor}
\usepackage{sidecap}
\usepackage[accsupp]{axessibility} 

\usepackage{stackengine}
\usepackage{xcolor}
\def\dottedcirc{\color{white}%
\stackinset{c}{}{c}{-5.pt}{--}{%
\stackinset{c}{}{c}{-5.pt}{--}{%
\stackinset{c}{}{c}{.1pt}{\rotatebox{90}{$-$}}{%
\stackinset{c}{}{c}{.1pt}{\rotatebox{45}{$-$}}{%
\stackinset{c}{}{c}{.1pt}{\rotatebox{-45}{$-$}}{%
\textcolor{red}{$\circ$}%
}%
}}}}\color{black}}
\usepackage{floatrow}

\usepackage[pagebackref=true,breaklinks=true,letterpaper=true,colorlinks,bookmarks=false]{hyperref}

\iccvfinalcopy 

\def\iccvPaperID{6485} 

\ificcvfinal\fi

\usepackage[capitalize]{cleveref}
\crefname{section}{Sec.}{Secs.}
\Crefname{section}{Section}{Sections}
\Crefname{table}{Table}{Tables}
\crefname{table}{Tab.}{Tabs.}

\newcommand\joonseok[1]{\textcolor{black}{#1}}

\newcommand\red[1]{\textcolor{red}{#1}}

\begin{document}

\newcommand{\propose}{POTR-3D}
\title{Towards Robust and Smooth 3D Multi-Person Pose Estimation
\\from Monocular Videos in the Wild}
%


\author{Sungchan Park \qquad Eunyi You \qquad Inhoe Lee \qquad Joonseok Lee\thanks{Corresponding author}\\
Seoul National University\\
{\tt\small \{warld2234, onlyou0416, inhoelee, joonseok\}@snu.ac.kr}}
\maketitle

\begin{abstract}
3D pose estimation is an invaluable task in computer vision with various practical applications. 
Especially, 3D pose estimation for multi-person from a monocular video (3DMPPE) is particularly challenging and is still largely uncharted, far from applying
to in-the-wild scenarios yet.
We pose three unresolved issues with the existing methods: lack of robustness on unseen views during training, vulnerability to occlusion, and severe jittering in the output.
As a remedy, we propose \propose, the first realization of a sequence-to-sequence 2D-to-3D lifting model for 3DMPPE,
powered by a novel geometry-aware data augmentation strategy, capable of generating unbounded data with a variety of views while caring about the ground plane and occlusions.
Through extensive experiments, we verify that the proposed model and data augmentation 
robustly generalizes to diverse unseen views, robustly recovers the poses against heavy occlusions, and reliably generates more natural and smoother outputs.
The effectiveness of our approach is verified not only by achieving the state-of-the-art performance on public benchmarks, but also by qualitative results on more challenging in-the-wild videos. Demo videos are available at \href{https://www.youtube.com/@potr3d}{https://www.youtube.com/@potr3d}.


\end{abstract}

\vspace{-0.4cm}
\section{Introduction}
\label{sec:intro}

3D pose estimation aims to reproduce the 3D coordinates of a person appearing in an untrimmed 2D video.
It has been extensively studied in literature with many real-world applications, \textit{e.g.}, sports~\cite{bridgeman2019multi}, healthcare~\cite{wu2020human}, games~\cite{ke2010real}, movies~\cite{alahari2013pose}, and video compression~\cite{facevid2vidnvidia}.
Instead of fully rendering 3D voxels, we narrow down the scope of our discussion to reconstructing a handful number of body keypoints (\textit{e.g.}, neck, knees, or ankles), which concisely represent dynamics of human motions in the real world.


Depending on the number of subjects, 3D pose estimation is categorized into 3D Single-Person Pose Estimation (3DSPPE) and 3D Multi-Person Pose Estimation (3DMPPE).
In this paper, we mainly tackle 3DMPPE, reproducing the 3D coordinates of body keypoints for everyone appearing in a video.
Unlike 3DSPPE that has been extensively studied and already being used for many applications, 3DMPPE is still largely uncharted whose models are hardly applicable to in-the-wild scenarios yet.
For this reason, we pose 3 unresolved issues with previous models.


First, existing models are not robustly applicable to unseen views (\textit{e.g.} unusual camera angle or distance).
Trained on a limited amount of data, most existing models perform well only on test examples captured under similar views, significantly underperforming when applied to unseen views.
Unfortunately, however, most 3D datasets provide a limited number of views (\textit{e.g.}, 4 for Human 3.6M~\cite{ionescu2013human3} or 14 for MPI-INF-3DHP~\cite{mono-3dhp2017}), recorded under limited conditions like the subjects' clothing or lighting due to the high cost of motion capturing (MoCap) equipment~\cite{gong2021poseaug}.
This practical restriction hinders learning a universally applicable model, failing to robustly generalize to in-the-wild videos.

Second, occlusion is another long-standing challenge that most existing models still suffer from.
Due to the invisible occluded keypoints, there is unavoidable ambiguity since there are multiple plausible answers for them.
Occlusion becomes a lot more severe when a person totally blocks another from the camera, making the model output inconsistent estimation throughout the frames.

Third, the existing methods often produce a sequence of 3D poses with severe jittering.
This is an undesirable byproduct of the models, not present in the training data.

In this paper, we propose a 3DMPPE model, called \propose, that works robustly and smoothly on in-the-wild videos with severe occlusion cases.
Our model is powered by a novel data augmentation strategy, which generates training examples by combining and adjusting existing single-person data, to avoid actual data collection that requires high cost.
For this augmentation, we decouple the core motion from pixel-level details, as creating a realistic 3D video at the voxel level is not a trivial task.
Thereby, we adopt the 2D-to-3D lifting approach, which first detects the 2D body keypoints using an off-the-shelf model and trains a model to lift them into the 3D space.
Benefiting from a more robust and generalizable 2D pose estimator, 2D-to-3D lifting approaches have been successfully applied to 3DSPPE~\cite{pavllo20193d,zhang2022mixste}.
Observing that we only need the body keypoints, not the pixel-level details, of the subjects for training, we can easily generate an unlimited number of 2D-3D pairs using given camera parameters under various conditions, \textit{e.g.}, containing arbitrary number of subjects under various views.
Translating and rotating the subjects as well as the ground plane itself, our augmentation strategy makes our model operate robustly on diverse views.




To alleviate the occlusion problem, we take a couple of remedies.
The main reason that existing models suffer from occlusion is that they process a single frame at a time (\textit{frame2frame}).
Following a previous work, MixSTE~\cite{zhang2022mixste}, which effectively process multiple frames at once (\textit{seq2seq}) for 3DSPPE,
our \propose\ adopts a similar Transformer-based 2D-to-3D structure, naturally extending it to multi-person.
Lifting the assumption that there is always a single person in the video, \propose\ tracks multiple people at the same time, equipped with an additional self-attention across multiple people appearing in the same frame.
We infer the depth and relative poses in a unified paradigm, helpful for a comprehensive understanding of the scene.
To the best of our knowledge, \propose\ is the first \textit{seq2seq} 2D-to-3D lifting method for 3DMPPE.
Alongside, we carefully design the augmentation method to reflect occlusion among people with a simple but novel volumetric model.
We generate training examples with heavy occlusion, where expected outputs of off-the-shelf models are realistically mimicked, and thus the model is expected to learn from these noisy examples how to confidently pick useful information out of confusing situations.



The \textit{seq2seq} approach also helps the model to reduce jittering, allowing the model to learn temporal dynamics by observing multiple frames at the same time.
In addition, we propose an additional loss based on MPJVE, which has been introduced to measure temporal consistency~\cite{pavllo20193d}, to further smoothen the prediction across the frames.


In summary, we propose \propose, the first realization of a \textit{seq2seq} 2D-to-3D lifting model for 3DMPPE, and devise a simple but effective data augmentation strategy, allowing us to generate an unlimited number of occlusion-aware augmented data with diverse views.
Putting them together, our overall methodology effectively tackles the aforementioned three challenges in 3DMPPE and adapts well to in-the-wild videos. Specifically,
\begin{itemize}
    \setlength{\itemsep}{0pt}
    \setlength{\parskip}{0pt}
    \item Our method robustly generalizes to a variety of views that are \textit{unseen during training}, overcoming the long-standing data scarcity challenge. 
    \item Our approach robustly recovers the poses in situations with \textit{heavy occlusion}.
    \item Our method produces a more \textit{natural and smoother} sequence of motion compared to existing methods.
\end{itemize}
\vspace{-0.2cm}
Trained on our augmented data, \propose\ outperforms existing methods both quantitatively on several representative benchmarks and qualitatively on in-the-wild videos.

\section{Related Work}
\label{sec:related}




3D Human pose estimation has been studied on single-view (monocular) or on multi-view images.
Seeing the scene only from one direction through a monocular camera, the single-view pose estimation is inherently challenging to reproduce the original 3D landscape.
Multi-view systems~\cite{kanade1997virtualized,hofmann2009multi,hofmann2012multi,joo2015panoptic,iskakov2019learnable,remelli2020lightweight,he2020epipolar,bultmann2021real,zhang2021direct} are developed to ease this problem.
In this paper, we focus on the monocular 3D human pose estimation, as we are particularly interested in in-the-wild videos captured without special setups.




\textbf {3D Single-Person Pose Estimation (3DSPPE).}
There are two directions tackling this task. The first type directly constructs 3D pose from a given 2D image or video end-to-end at pixel-level~\cite{pavlakoscoarsetofine, xiaosuncompositional}.
Another direction, the 2D-to-3D lifting, treats only a few human body keypoints instead of pixel-level details, leveraging off-the-shelf 2D pose estimator trained on larger data.
VideoPose3D~\cite{pavllo20193d} performs sequence-based 2D-to-3D lifting for 3DSPPE using dilated convolution.
Some recent works~\cite{zhao2019semantic,liu2020comprehensive,ci2019optimizing} apply Graph Neural Networks~\cite{kipf2016semi} to 2D-to-3D lifting.
PoseFormer~\cite{poseformerzheng20213d} utilizes a Transformer for 3DSPPE to capture the spatio-temporal dependency, referring to a sequence of 2D single-person pose from multiple frames at once to estimate the pose of the central frame (\textit{seq2frame}).
It achieves competent performance, but redundant computation is known as a drawback since large amount of sequences overlaps to infer 3D poses for all frames. 
MixSTE~\cite{zhang2022mixste} further extends it to reconstruct for all frames at once, better modeling sequence coherence and enhancing efficiency.

\textbf {3D Multi-Person Pose Estimation (3DMPPE).}
Lifting the assumption that only a single person exists in the entire video, 3DMPPE gets far more challenging.
In addition to the relative 3D position of body keypoints for each individual,
depth of all persons in the scene also should be predicted in 3DMPPE, as geometric relationship between them does matter.
Also, each individual needs to be identified across frames.


Similarly to the 3DSPPE, 3D poses may be constructed end-to-end.
Moon \textit{et al.}~\cite{Moon_2019_ICCV_3DMPPE} directly estimates each individual's depth assuming a typical size of 3D bounding box.
However, this approach fails when the pose of a subject significantly varies.
BEV~\cite{BEV} explicitly estimates a bird's-eye-view to better model the inherent 3D geometry.
It also infers the height of each individual by age estimation.

2D-to-3D lifting is also actively proposed.
Ugrinovic \textit{et al.}~\cite{SizeDepth} considers the common ground plane, where appearing people are standing on, to help depth disambiguation.
They firstly estimate 2D pose and SMPL~\cite{SMPL2015} parameters of each person, and lift them to 3D with the ground plane constraint and reprojection consistency.
VirtualPose~\cite{VirtualPose2022} extracts 2D representation from heat maps of keypoints and lifts them to 3D, trained on a synthetic paired dataset of 2D representation and 3D pose.

However, all aforementioned methods are \textit{frame2frame} approaches, vulnerable to occlusion, which is the main challenge in  3DMPPE.
Multiple frames may be considered simultaneously to resolve occlusion, but end-to-end approaches are limited to enlarge the receptive field due to heavy computation (\textit{e.g.}, 3D convolution).
2D-to-3D lifting approach is potentially suitable, but no previous work has tried to combine temporal modeling with it, to the best of our knowledge.
Here, Each subject should be re-identified, as most off-the-shelf 2D pose estimator operates in \textit{frame2frame} manner.
By using an additional off-the-shelf 2D pose tracking model with some adaptation and geometry-aware data augmentation, we tackle the challenges in 3DMPPE with a \textit{seq2seq} approach.

\textbf{Data Augmentation for 3D Pose Estimation.}
Since 3D pose annotation is expensive to collect, limited training data is another critical challenge that overfits a model.
For this, several data augmentation methods have been proposed.

The 2D-to-3D lifting approach allows unbounded data augmentation of 2D-3D pose pairs by decoupling the 3D pose estimation problem into 2D pose estimation and 2D-to-3D lifting.
Most 2D-to-3D lifting methods apply horizontal flipping~\cite{pavllo20193d,poseformerzheng20213d,zhang2022mixste,VirtualPose2022}. PoseAug~\cite{gong2021poseaug} suggests adaptive differentiable data augmentation for 3DSPPE which adjusts bone angle, bone length, and body orientation.

Data augmentation for 3DMPPE gets more complicated, since geometric relationship (\textit{e.g.} distance) among individuals should be additionally considered.
VirtualPose~\cite{VirtualPose2022} proposes random translation and rotation of each person, and trains the lifting model solely with the augmented data.
This work shares the motivation with ours, but it does not care about the augmentation of the ground plane or camera view, and it does not simulate occlusion cases.
Our augmentation method, on the other hand, explicitly considers the ground plane to make the result feasible and to help disambiguation of depth~\cite{SizeDepth}.
Also, we further propose to augment the ground plane itself for robustness on various views.
In addition, we specially treat occluded keypoints to simulate actual occlusion cases.

\textbf{Depth Estimation on a Monocular Image}.
It is in nature an ill-posed problem to estimate the depth from a monocular input, since the size of an object is coupled with it.
For instance, if an object gets 2$\times$ larger and moves away from the camera by 2$\times$ distance, the object will still look the same in the projected 2D image.
Thus, an additional clue is needed to estimate the distance, \textit{e.g.}, typical size of well-known objects like a person.
Assuming all people in the scene roughly have a similar height, one can estimate a plausible depth.
Some previous works, \textit{e.g.}, BEV~\cite{BEV}, try to disambiguate heights of people by inferring age.
We admit that such an attempt may somehow beneficial, but in this paper, we do not explicitly infer the size of people.
The model might implicitly learn how to estimate depth of a person with a usual height observed in the training data.
This might result in a sub-optimal result in some complicated cases, \textit{e.g.}, a scene with both adults and children. We leave this disambiguation as an interesting future work to concentrate more on aforementioned issues.


\begin{figure*}
    \includegraphics[width=\linewidth]{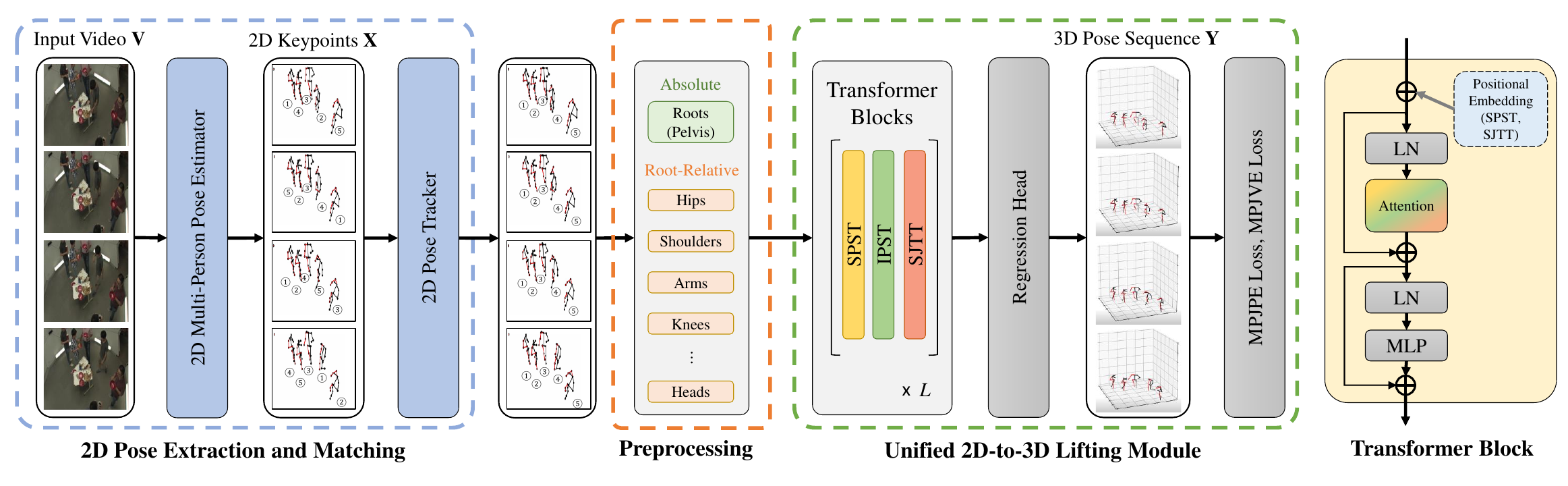}
    \caption{
        \textbf{Overview of the \propose.}
        The input video is converted to 2D keypoints, followed by 2D-to-3D lifting, composed of stacked three types of Transformers (SPST, IPST, SJTT).}
    \label{fig:overview}
\end{figure*}

\section{Preliminaries}

\textbf{Problem Formulation.}
In the 3D Multi-person Pose Estimation (3DMPPE) problem, the input consists of a video $\mathbf{V} = [\mathbf{v}_1, ..., \mathbf{v}_T]$ of $T$ frames, where each frame is $\mathbf{v}_t \in \mathbb{R}^{H \times W \times 3}$ and (up to) $N$ persons may appear in the video.
The task is locating a predefined set of $K$ human body keypoints (\textit{e.g.}, neck, ankles, or knees; see Fig.~\ref{fig:occlusion} for an example) in the 3D space for all persons appearing in the video in every frame.
The body keypoints in the 2D image space are denoted by $\mathbf{X} \in \mathbb{R}^{T \times N \times K \times 2}$, and the output $\mathbf{Y} \in \mathbb{R}^{T \times N \times K \times 3}$ specifies the 3D coordinates of each body keypoint for all $N$ people across $T$ frames.
We follow the common assumption that the camera is static.


\textbf{Notations.}
For convenience, we define a common notation for 2D and 3D points throughout the paper.
Let us denote a 2D point $\mathbf{X}_{t,i,k} \in \mathbb{R}^2$ as $(u, v)$, where $u \in \{0, ..., H - 1\}$ and $v \in \{0, ..., W - 1\}$ is the vertical and horizontal coordinate in the image, respectively.
Similarly, we denote a 3D point within the camera coordinate $\mathbf{Y}_{t,i,k} \in \mathbb{R}^3$ as $(x, y, z)$, where $x$ and $y$ are the coordinates through the two directions parallel to the projected 2D image, and $z$ is the depth from the camera.


\textbf{Data Preprocessing.}
We adjust the input in two ways, following common practice.
First, we specially treat a keypoint called a root joint (typically pelvis, the body center; denoted by $\mathbf{Y}_{t,i,1} \in \mathbb{R}^3$) for a person $i$ at frame $t$.
The ground truth for this point is given by $(u, v, z)$, where $(u, v)$ is the true 2D coordinate of the root joint and $z$ is its depth.
For root joints, the model estimates $(\hat{u}, \hat{v}, \hat{z})$, the absolute values for $(u, v, z)$.
(Note that $\hat{u} \approx u$ and $\hat{v} \approx v$ but they are still estimated to compensate for imperfect 2D pose estimation by the off-the-shelf model.)
Other regular joints, $\mathbf{Y}_{t,i,k} \in \mathbb{R}^3$ for $k = 2, ..., K$, are represented as the relative difference from the corresponding root joint, $\mathbf{Y}_{t,i,1}$.

Second, we normalize the ground truth depth of the root joints by the camera focal length.\footnote{Note that the depts of other joints are not normalized.}
When a 2D pose is mapped to the 3D space, the depth of each 2D keypoint towards the direction of projection needs to be estimated.
Since the estimated depth would be proportional to the camera focal length used at training, we normalize the ground truth depth $z$ by the focal length, following common practice.
That is, we use $\Bar{z} = z / f$, where $f$ is the camera focal length.
In this way, our model operates independently of the camera focal length.

\section{The \propose\ Model}
\label{sec:method}

The overall model workflow is depicted in \cref{fig:overview}.
First, the input frames $\mathbf{V}$ are converted to a sequence of 2D keypoints by an off-the-shelf model (Sec.~\ref{sec:method:preprocess}).
Then, they are lifted into the 3D space (Sec.~\ref{sec:method:2dto3d}).

\subsection{2D Pose Extraction and Matching}
\label{sec:method:preprocess}

Given an input video $\mathbf{V} \in \mathbb{R}^{T \times H \times W \times 3}$, we first extract the 2D coordinates $\mathbf{X} \in \mathbb{R}^{T \times N \times K \times 2}$ of the (up to) $N$ persons appearing in the video, where $T$ is the number of frames, and $K$ is the number of body keypoints, determined by the dataset.
Since we treat multiple people in the video, each individual should be re-identified across all frames.
That is, the second index of $\mathbf{X}$ and $\mathbf{Y}$ must be consistent for each individual across all frames. 
Any off-the-shelf 2D multi-person pose estimator and a tracking model can be adopted for each, and we use HRNet~\cite{sun2019deep} and ByteTrack~\cite{zhang2022bytetrack}, respectively.
Note that these steps need to be done only at testing, since we train our model on augmented videos, where the 2D coordinates can be computed from the ground truth 3D poses and camera parameters (see Sec.~\ref{sec:augmentation}).



\subsection{Unified 2D-to-3D Lifting Module}
\label{sec:method:2dto3d}

From the input $\mathbf{X}$, this module lifts it to the 3D coordinates, $\mathbf{Y} \in \mathbb{R}^{T \times N \times K \times 3}$.
To effectively comprehend the spatio-temporal geometric context, we adapt encoder of \cite{zhang2022mixste}.
The coordinates of each 2D keypoints $\mathbf{X}_{t,i,k}$, at a specific frame $t \in \{1, ..., T\}$ for a specific person $i \in \{1, ..., N\}$ and body keypoint $k \in \{1, ..., K\}$, is linearly mapped to a $D$-dimensional token embedding.
Thus, the input is now converted to a sequence of $T \times N \times K$ tokens in $\mathbb{R}^{D}$, and let us denote these tokens as $\mathbf{Z}^{(0)} \in \mathbb{R}^{T \times N \times K \times D}$. Here, in contrast to most existing methods \cite{Moon_2019_ICCV_3DMPPE,BEV,VirtualPose2022}, which separately process the root and regular joints,
we treat them with a single unified model.
This unified model is not just simpler, but also enables more comprehensive estimation of depth and pose by allowing attention across them.

\begin{figure}
    \centering
    \includegraphics[width=1.0\linewidth]{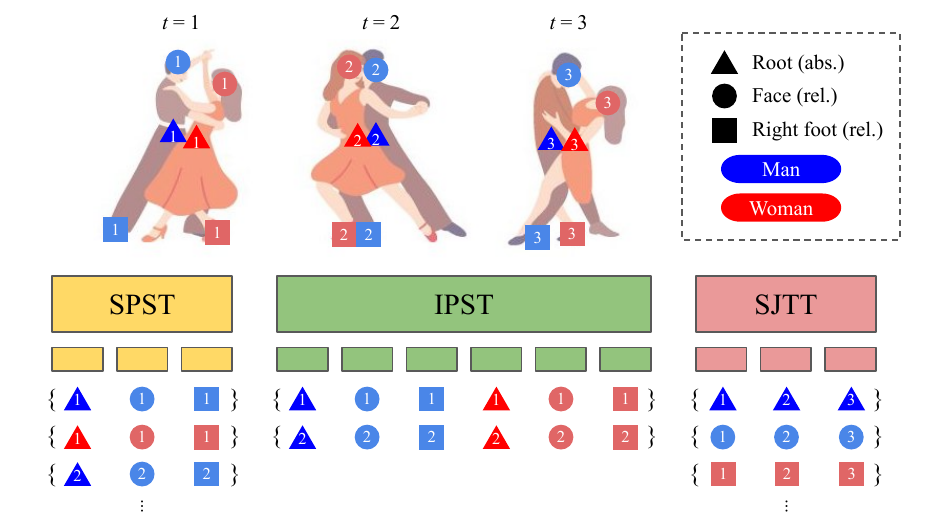}
    \caption{Illustration of 2D-to-3D Lifting Transformers.}
    \label{fig:2d_to_3d_illustration}
    \vspace{-0.3cm}
\end{figure}

They are fed into the repeated 3 types of Transformers, where each of them is designed to model a specific relationship between different body keypoints.
This factorization is also beneficial to reduce the quadratic computation cost of the attention mechanism.
In addition to the two Transformers in \cite{zhang2022mixste}, SPST for modeling spatial and SJTT for temporal relationships, we propose an additional Transformer (IPST) designed to learn inter-person relationships among all body keypoints.
The role of each Transformer is illustrated in \cref{fig:2d_to_3d_illustration} and detailed below.
The input $\mathbf{Z}^{(\ell-1)} \in \mathbb{R}^{T \times N \times K \times D}$ at each layer $\ell$ goes through SPST, IPST, and SJTT to contextualize within the sequence, and outputs the same sized tensor $\mathbf{Z}^{(\ell)} \in \mathbb{R}^{T \times N \times K \times D}$.



\textbf{Single Person Spatial Transformer (SPST).}
Located at the first stage of each layer $\ell$, SPST learns spatial correlation of each person's joints in each frame.
From the input $\mathcal{X} \in \mathbb{R}^{T \times N \times K \times D}$, SPST takes $K$ tokens of size $D$ corresponding to $\mathcal{X}_{t, i} \in \mathbb{R}^{K \times D}$ for $t \in \{1, ..., T\}$ and $i \in \{1, ..., N\}$, separately at a time.
In other words, SPST takes $K$ different body keypoints belonging to a same person $i$ at a specific frame $t$.
The output $\mathcal{Y} \in \mathbb{R}^{T \times N \times K \times D}$ has the same shape, where each token $\mathcal{Y}_{t,i,k}$ is a transformed one by contextualizing across other tokens belonging to the same person.


\textbf{Inter-Person Spatial Transformer (IPST).}
After SPST, IPST learns correlation among multiple individuals in the same frame.
Through this, the model learns spatial inter-personal relationship in the scene.
\joonseok{IPST is newly designed to extend the previous 3DSPPE model to 3DMPPE.}
More formally, IPST takes $N \times K$ tokens of size $D$ as input at a time; that is, given the input $\mathcal{X} \in \mathbb{R}^{T \times N \times K \times D}$, all $N \times K$ tokens in the frame $\mathcal{X}_{t} \in \mathbb{R}^{N \times K \times D}$ are fed into IPST, contextulize from each other, and are transformed to the output tokens $\mathcal{Y}_t$.
This process is separately performed for each frame at $t = 1, ..., T$.
After IPST, each token is knowledgeable about other people in the same scene.

This is a novel part of our method to deal with multiple individuals.
Different from the TDBU-Net~\cite{Cheng_2021_CVPR}, which uses a discriminator to validate relationship between subjects at the final stage, IPST progressively enhances its contextual understanding on inter-personal relationship by iterative attentions.

\textbf{Single Joint Temporal Transformer (SJTT).}
From the input $\mathcal{X} \in \mathbb{R}^{T \times N \times K \times D}$, we create $N \times K$ input sequences of length $T$, corresponding to $\mathcal{X}_{\cdot,i,k} \in \mathbb{R}^{T \times D}$ for $i = 1, ..., N$ and $k = 1, ..., K$.
Each sequence is fed into SJTT, temporally contextualizing each token in the sequence, and the transformed output tokens $\mathcal{Y}_{\cdot,i,k}$ are returned.
Completing all $N \times K$ sequences, the transformed sequence $\mathcal{Y} \in \mathbb{R}^{T \times N \times K \times D}$ is output as the result of the $\ell$-th layer of our 2D-to-3D lifting module, $\mathbf{Z}^{(\ell)}$.

These 3 blocks constitute a single layer of our 2D-to-3D lifting module, and multiple such layers are stacked.
A learnable positional encoding is added to each token at the first layer ($\ell= 1$) of SPST and SJTT.
No positional encoding is added for IPST, since there is no natural ordering between multiple individuals in a video.

\textbf{Regression Head.}
After repeating $L$ layers of \{SPST, IPST, SJTT\}, we get the output tokens for all body keypoints, $\mathbf{Z}^{(L)} \in \mathbb{R}^{T \times N \times K \times D}$.
This is fed into a regression head, composed of a multilayer perceptron (MLP).
It maps each body keypoint embedding in $\mathbf{Z}^{(L)}$ to the corresponding 3D coordinates, $\mathbf{\hat{Y}} \in \mathbb{R}^{T \times N \times K \times 3}$.

\subsection{Training Objectives}
\label{sec:method:objective}

Given the predicted $\mathbf{\hat{Y}} \in \mathbb{R}^{T \times N \times K \times 3}$ and ground truth $\mathbf{Y} \in \mathbb{R}^{T \times N \times K \times 3}$, we minimize the following two losses.

\textbf{Mean per Joint Position Error (MPJPE) Loss} is the mean $L_2$ distance between the prediction and the target:
\begin{equation}
    \mathcal{L}_\text{MPJPE} = \frac{1}{TNK} \sum_{t=1}^T \sum_{i=1}^N \sum_{k=1}^{K} \| \mathbf{\hat{Y}}_{t,i,k} - \mathbf{Y}_{t,i,k} \|_2.
    \label{eq:mpjpe}
\end{equation}



\textbf{Mean per Joint Velocity Error (MPJVE) Loss}~\cite{zhang2022mixste} is the mean $L_2$ distance between the first derivatives of the prediction and the target with regards to time, measuring smoothness of the predicted sequence:
\begin{equation}
    \mathcal{L}_\text{MPJVE} = \frac{1}{TNK} \sum_{t=1}^T \sum_{i=1}^N \sum_{k=1}^{K} \left\| \frac{\partial \mathbf{\hat{Y}}_{t,i,k}}{\partial t} - \frac{\partial \mathbf{Y}_{t,i,k}}{\partial t} \right\|_2.
    \label{eq:mpjve}
\end{equation}


The overall loss $\mathcal{L}$ is a weighted sum of the two losses; that is,
$\mathcal{L} = \mathcal{L}_\text{MPJPE} + \lambda \cdot \mathcal{L}_\text{MPJVE}$,
where $\lambda$ controls the relative importance between them.
Optionally, different weights can be applied to the root joints and others.

\subsection{Inference and Scalability}

To perform inference on a long video, we first track every subject throughout the whole frames, then the tracked 2D keypoints are lifted to 3D at the regularly-split clip level.
The final result is obtained by concatenating the clip-level lifting.
In this way, our approach is scalable for a long video without extremely growing computational cost.
\section{Geometry-Aware Data Augmentation}
\label{sec:augmentation}

Being free from the pixel-level details, we can freely augment the training data as proposed below, making \propose\ robust on diverse inputs, resolving the data scarcity issue for this task.


Specifically, we take $N$ samples $\mathbf{Y}^{(i)} \in \mathbb{R}^{T \times K \times 3}$ captured by a fixed camera from a single-person dataset, where $i = 1, ..., N$.
We simply overlay them onto a single tensor $\mathbf{Y}\in \mathbb{R}^{T \times N \times K \times 3}$, and project it to the 2D space by applying the perspective camera model, producing $\mathbf{X} \in \mathbb{R}^{T \times N \times K \times 2}$.
A point $(x, y, z)$ in the 3D space is mapped to $(u, v)$ by
\begin{equation}
    \left[
        \begin{matrix}
            u \\ v \\ 1
        \end{matrix}
    \right] \approx
    \left[
        \begin{matrix}
            f_u & 0 & c_u \\
            0 & f_v & c_v \\
            0 & 0 & 1
        \end{matrix}
    \right]
    \left[
        \begin{matrix}
            x \\ y \\ z
        \end{matrix}
    \right],
\end{equation}
where $f_u, f_v$ are the focal lengths, and $c_u, c_v$ are the principal point in the 2D image coordinates.
This $(\mathbf{X}, \mathbf{Y})$ is an augmented 3DMPPE training example, and repeating this with different samples will infinitely generate new ones.

Furthermore, we consider additional augmentation on the trajectories, \textit{e.g.}, randomly translating or rotating them, to introduce additional randomness and fully take advantage of existing data.
However, there are a few geometric factors to consider: the ground plane and potential occlusion.

\subsection{Ground-plane-aware Augmentations}
\label{sec:augmentation:ground_plane}

\begin{figure}
    \centering
    \includegraphics[width=\linewidth]{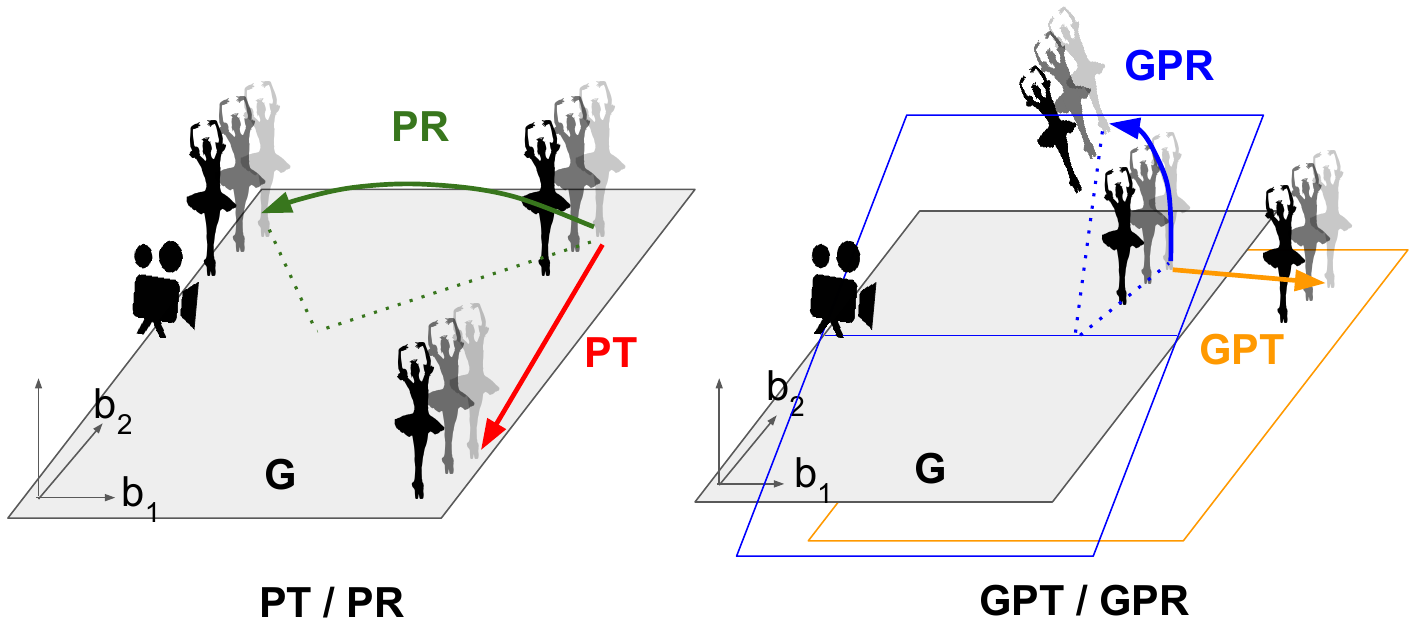}
    \caption{
        Illustration of the proposed data augmentation methods.}
    \label{fig:Augmentation}
\end{figure}


Although translating or rotating a trajectory in 3D space sounds trivial, most natural scenes do not fully use the three degrees of freedom, because of an obvious fact that people usually stand on the ground.
Geometrically, people in a video share the common ground plane, with a few exceptions like a swimming video.
As feet generally touch the ground, we estimate the ground plane by collecting feet coordinates from all frames captured by a fixed camera and fit them with linear regression, producing a 2D linear manifold $G$ within the 3D space.
We choose its two basis vectors, $\{\mathbf{b}_1, \mathbf{b}_2\}$, perpendicular to the normal vector of $G$.

We generate abundant sequences mimicking various multi-person and camera movements by combining 4 augmentation methods, illustrated in \cref{fig:Augmentation}:
\begin{itemize}
    \setlength{\itemsep}{0pt}
    \setlength{\parskip}{0pt}
    \item \textbf{Person Translation (PT)}: The target person is translated randomly along the basis $\{\textbf{b}_1, \textbf{b}_2\}$ on the ground plane.
    Each individual moves by a random amount of displacement $(\alpha, \beta)$ towards $\{\textbf{b}_1, \textbf{b}_2\}$.
    
    \item \textbf{Person Rotation (PR)}: 
    The person is rotated by an angle $\theta$ across the entire sequence to preserve natural movement, with respect to the normal vector of the ground plane about the origin at the averaged root joint over time of all subjects.

    \item \textbf{Ground Plane Translation (GPT)}: 
    The entire ground plane is shifted through the depth ($z$) axis by 
    $\gamma$, towards ($\gamma < 0$) or away from the camera. Applied to the ground plane, it affects everyone homogeneously.

    \item \textbf{Ground Plane Rotation (GPR)}:
    The entire ground plane is rotated by 
    $\varphi$ with respect to $\mathbf{b}_1$, whose direction is more parallel to the $x$-direction of camera ($\sim$pitch rotation). This is to generate vertically diverse views, which are challenging to create with the other 3 augmentations.
\end{itemize}





\subsection{Reflecting Occlusions}
\label{sec:augmentation:occlusion}

\begin{figure}
  \centering
  \includegraphics[width=3.2cm,height=2.3cm]{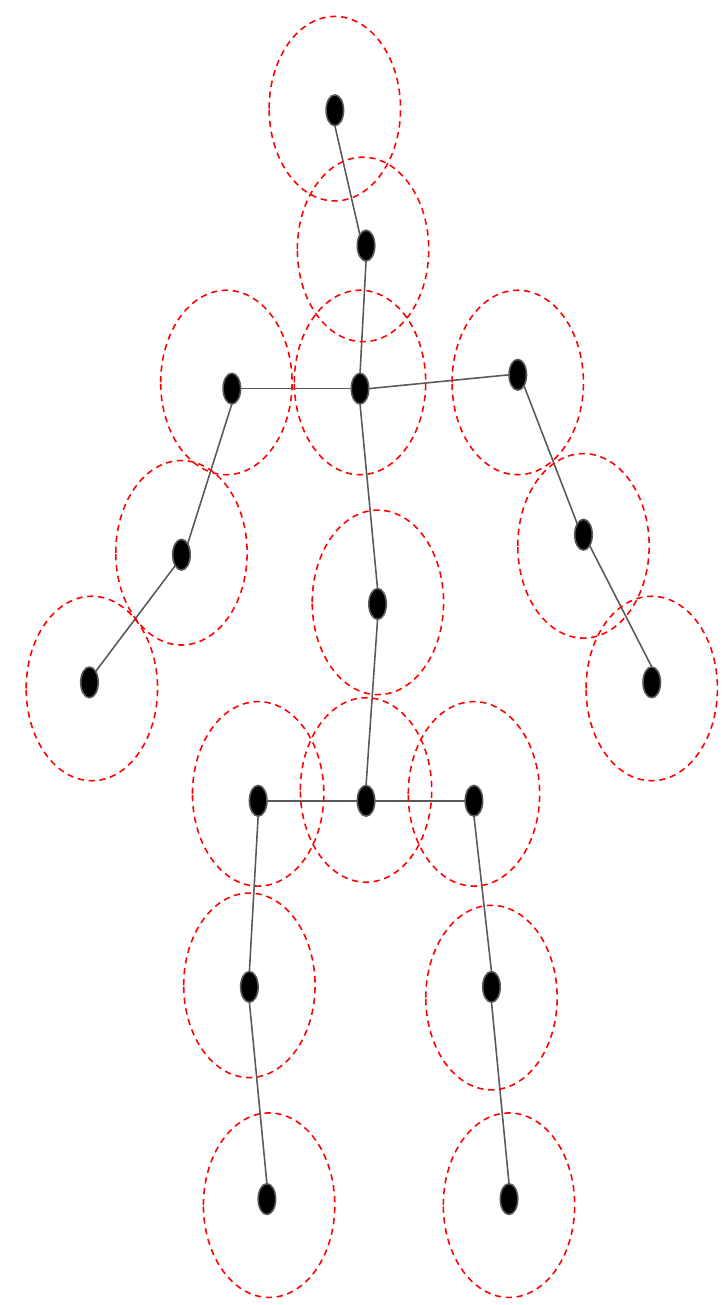}
  \caption{\textbf{Our volume representation of a person to simulate occlusion.} \textbf{$\bullet$} indicates the keypoints, and \textbf{\red{$\bigcirc$}} indicates the 3D balls of radius $R$ (\textit{i.e.} $14$cm).}
  \label{fig:occlusion}
\end{figure}

{ 
}

2D pose estimators often suffer from occlusion.
In order for the model to learn how to pick useful and confident information without being confused and be robust against occlusion, we simulate occlusion on our data augmentation.


As the human body has some volume, two body parts (either from the same person or from different ones) may occlude if the two keypoints are projected close enough, even though they do not exactly coincide.
From this observation, we propose a simple volume representation of a person, illustrated in \cref{fig:occlusion}.
The volume of each body part is modeled as a same-sized 3D ball centered at the corresponding keypoint.
Once projected to the 2D plane, a ball becomes a circle.
The circles are considered to overlap if the distance between their centers is shorter than the larger one's radius.
Then, the one with the smaller radius, which is farther away, is regarded as occluded.
If a keypoint is occluded, we either perturb it with some noise or drop it out, simulating the case where keypoints predicted by the off-the-shelf model with low confidence are dropped at inference.
Dropped keypoints are represented as a learned mask token.


Although there exist more sophisticated methods like a cylinder man model~\cite{occawarenet}, computation overhead of judging overlaps also gets more severe.
Our sphere model is simpler and computationally cheaper. Especially on 3DMPPE, we expect less gap between the two, as the size of targets is relatively smaller than usual distance between them.





Once a training example is generated, its validity needs to be checked, \textit{e.g.}, positivity of the depth $z$ for all targets, validity of the randomly chosen displacements, and more.
See \cref{sec:supp:augmentation} for more details.




\section{Experiments}
\label{sec:exp}

\subsection{Experimental Settings}
\label{sec:exp:setting}

\noindent
\textbf{Datasets.}
MuPoTS-3D~\cite{singleshotmultiperson2018} is a representative dataset for monocular 3DMPPE, composed of 20 few-second-long sequences with 2--3 people interacting with each other.
Since this data is made only for evaluation purpose, MuCo-3DHP~\cite{singleshotmultiperson2018} is widely paired with it for training.
MuCo-3DHP is artificially composited from a single-person dataset MPI-INF-3DHP~\cite{mono-3dhp2017}, which contains 8 subjects' various motions captured from 14 different cameras.
We follow the \textit{all annotated poses} evaluation setting, as opposed to the \emph{detected-only}.

CMU Panoptic \cite{joo2015panoptic} is another popular 3D multi-person dataset,
containing 60 hours of video with 3D poses and tracking information captured by multiple cameras.
Following \cite{zanfir2018}, we use video sequences of camera 16 and 30 for both training and testing.
This training set consists of sequences with \textit{Haggling}, \textit{Mafia}, and \textit{Ultimatum}, and the test set consists of sequences with an additional activity, \textit{Pizza}.

We train our model on a synthesized set using the proposed augmentation in \cref{sec:augmentation}.
We use MPI-INF-3DHP as the source of augmentation for MuPoTS-3D experiment. For CMU-Panoptic, we augment on its training partition.

\vspace{0.1cm} \noindent
\textbf{Evaluation Metrics.}
Following the conventional setting, we measure Percentage of Correct Keypoints (PCK; $\%$) for MuPoTS-3D, and MPJPE(mm) for CMU-Panoptic.
In addition, we further consider MPJVE(mm) to measure smoothness, which is barely investigated in 3DMPPE for both datasets.
While measuring PCK, a keypoint prediction is regarded as correct if the $L_2$ distance between the prediction and the ground truth is within a given threshold $\tau=150$mm.
We report PCK metrics with (PCK$_\text{rel}$) and without (PCK$_\text{abs}$) the root alignment. Higher PCK indicates better performance.
MPJPE (Eq.~\eqref{eq:mpjpe}) measures accuracy of the prediction, while MPJVE (Eq.~\eqref{eq:mpjve}) measures smoothness or consistency over frames.
Lower MPJPE and MPJVE indicate better performance.



\vspace{0.1cm} \noindent
\textbf{Competing Models.}
We compare \propose\ with 6 baselines: VirtualPose~\cite{VirtualPose2022}, BEV~\cite{BEV}, SingleStage~\cite{nie2019single}, SMAP~\cite{zhen2020smap}, SDMPPE~\cite{Moon_2019_ICCV_3DMPPE}, and MubyNet~\cite{zanfir2018deep}.
Models that additionally use ground truth at testing~\cite{Cheng_2021_CVPR,hmor_hierarchical} are still listed with remarks.

\vspace{0.1cm} \noindent
\textbf{Implementation Details.} 
The input 2D pose sequences are obtained by fine-tuned HRNet \cite{sun2019deep} and ByteTrack~\cite{zhang2022bytetrack}.
More details are in \cref{sec:supp:details}.
These off-the-shelf models are used only at testing, since we solely train on the augmented data.
If the number of tracked or augmented persons is less than $N$, we pad zeros.
We use Adam optimizer \cite{kingma2014adam} with batch size of 16, dropout rate of 0.1, and GELU activation.
We use two NVIDIA A6000 GPUs for training.




\subsection{Quantitative Comparison}
\label{sec:result:quant}

\begin{table}
\resizebox{\linewidth}{!}{ 
    \setlength{\tabcolsep}{2pt}
    \centering
    \begin{tabular}{l|cccc|cccc}
    \toprule
    Test set & \multicolumn{4}{c|}{Entire test set} & \multicolumn{4}{c}{Occlusion subset} \\
    Metric & PCK$_\text{rel}^\uparrow$ & PCK$_\text{abs}^\uparrow$ & MPJVE$_\text{rel}^\downarrow$ & MPJVE$_\text{abs}^\downarrow$ & PCK$_\text{rel}^\uparrow$ & PCK$_\text{abs}^\uparrow$ & MPJVE$_\text{rel}^\downarrow$ & MPJVE$_\text{abs}^\downarrow$ \\ 
    \midrule
    TDBU-Net~\cite{Cheng_2021_CVPR}$^\dag$     & 89.6      & 48.0            & 29.6  & 43.3  & 84.2     & 41.7            &  35.6  & 54.6 \\
    \midrule
    3DMPPE~\cite{Moon_2019_ICCV_3DMPPE} & {81.8}      & 31.5      & {24.0}  & {120.4}  & 76.7      & 30.8      & 28.0  & 137.0\\
    SMAP~\cite{zhen2020smap}            & 73.5      & 35.2      & --  & -- & 64.4      & 31.3     & --  & -- \\
    SingleStage~\cite{Jin_2022_CVPR}    & 80.9      & 39.3      & --  & -- & --      & --      & --  & -- \\
    BEV~\cite{BEV}  &  70.2     & --      & --  & -- &  --     & --      & --  & -- \\
    VirtualPose~\cite{VirtualPose2022}  &  72.3$^*$     & {44.0}      &  23.1  & 41.5  &  69.0$^*$     & 36.4      & 23.4  & 47.2  \\
    \propose\ (Ours)                                & \bf{83.7} & \bf{50.9} & \bf{10.9}  & \bf{16.3} & \bf{82.1} & \bf{47.2} & \textbf{12.2}  & \textbf{17.3} \\
    \bottomrule

    \end{tabular}
} 
    \caption{
        \textbf{Comparison on MuPoTS-3D.} The best scores are marked in boldface. TDBU-Net~\cite{Cheng_2021_CVPR}$^\dag$ is not comparable with other methods as it uses GT at testing. (*indicates our reproduction.) 
    } 
    \label{tab:result_mupots}
\end{table}

\noindent
\textbf{MuPoTS-3D.}
\cref{tab:result_mupots} compares 3DMPPE peformance in four metrics, on the entire MuPoTS-3D test set and on its subset of 5 videos (TS 2, 13, 14, 18, 20) with most severe occlusion.
Ours achieves the best scores on both test sets, significantly outperforming all baselines in all metrics.
The performance gap is larger on the occlusion subset.
\propose\ also outperforms on sequences with some unusual distance from the camera (\textit{e.g.}, TS6, 13).
These results indicate the effectiveness of our method and the augmentation.
(See \cref{tab:result_ts} in \cref{sec:supp:quant} for performance on individual videos.)


As MPJVE has not been reported in previous works, we reproduce it only for methods open-sourced.
\propose\ significantly outperforms other methods in MPJVE.
This is expected, since \propose\ is directly optimized over the same loss.
In fact, most baseline models cannot be optimized over MPJVE by nature and have overlooked it, since they operate in \textit{frame2frame}.
However, the MPJVE that our model achieves is still significant, considering that it outperforms the other \textit{seq2seq} baseline \cite{Cheng_2021_CVPR} which uses GT pose and depth at inference.





\begin{table}
\resizebox{\linewidth}{!}{ 
    \centering
    \begin{tabular}{l|cccc|c|c}
    \toprule
    Metric                              & \multicolumn{5}{c|}{MPJPE$_\text{rel}^\downarrow$}   & MPJVE$_\text{rel}^\downarrow$ \\
    \midrule
    Sequence                                & \textit{Haggling}  & \textit{Mafia} & \textit{Ultimatum} & \textit{Pizza} & Avg. & Avg. \\ 
    \midrule
    HMOR~\cite{hmor_hierarchical}$^\dag$
        & 50.9
        & 50.5
        & 50.7
        & 68.2
        & 55.1
        & -- \\
    \midrule
    MubyNet~\cite{zanfir2018deep}
        & 72.4
        & 78.8
        & 66.8
        & 94.3
        & 78.1
        & -- \\
    SMAP~\cite{zhen2020smap} 
        & 63.1
        & 60.3
        & 56.6
        & 67.1
        & 61.8
        & -- \\
    BEV~\cite{BEV} 
        & 90.7
        & 103.7
        & 113.1
        & 125.2
        & 108.2
        & -- \\
    VirtualPose~\cite{VirtualPose2022}  
        & \textbf{54.1}
        & 61.6
        & \textbf{54.6}
        & 65.4
        & 58.9
        & 13.7 \\

    \propose\ (Ours)
        & 60.0
        & \textbf{57.0}
        & 55.5
        & \textbf{58.9}
        & \textbf{57.8}
        & \textbf{3.9} \\
    \bottomrule
    \end{tabular}
} 
\caption{
    \textbf{Comparison on CMU Panoptic.}
    Models are trained on \{\textit{Haggling}, \textit{Mafia}, \textit{Ultimatum}\}, and generalized to \textit{Pizza}. The best scores are marked in boldface. HMOR~\cite{hmor_hierarchical}$^\dag$ is not comparable with other methods as it uses GT depth at testing. 
} 
\label{tab:result_panotic}
\end{table}

\vspace{0.1cm} \noindent
\textbf{CMU-Panoptic.}
As seen in \cref{tab:result_panotic}, \propose\ also achieves the state-of-the-art performance on CMU-Panoptic, $1.1$mm or $1.9\%$ leading the strongest baseline~\cite{VirtualPose2022}.
In contrast to MuPoTS-3D, CMU-Panoptic contains videos with a denser crowd of 3--8 people, making the tracking more challenging.
The result indicates that \propose\ operates well even in this challenging situation.
Also, \propose\ achieves significantly higher performance than others on the \textit{Pizza} sequence unseen at training, with $5.7$mm or $9.9\%$ gain, verifying its superior generalizability.

\begin{table}
  \centering
  \resizebox{0.54\linewidth}{!}{

  \begin{tabular}{l|cc}
    \toprule
    
     Model
     & PCK$_\text{rel}^\uparrow$ & MPJVE$_\text{rel}^\downarrow$  \\ 
    \midrule
    
    MixSTE  & 57.3 & 11.8 \\
    VideoPose3D & 52.8 & 11.4 \\

    \midrule
    
    \propose\ (Ours) & \textbf{83.7}  & \textbf{10.9}  \\
    
    \bottomrule
  \end{tabular}
  } 
  \caption{\centering Comparison with 3DSPPE methods on MuPoTS-3D}
  \label{tab:result_3dsppe_abl}
\end{table}

\vspace{0.1cm} \noindent
\textbf{Comparison with 3DSPPE Models.}
We also compare ours with 3DSPPE methods~\cite{zhang2022mixste, pavllo20193d} on MuPoTS-3D, by lifting individual tracklets using each model and then aggregating them.
As these models do not estimate the depth, assuming only a single person to exist, only relative metrics could be used to measure the performance.
As shown in \cref{tab:result_3dsppe_abl}, \propose\ outperforms 3DSPPE models by a large margin, indicating the importance of simultaneous reasoning of multiple persons for disambiguation.

\subsection{Ablation Study}
\label{sec:exp:ablation}

\noindent
\textbf{Model Components.}
Our 2D-to-3D lifting module consists of three types of Transformers, SPST, IPST, and SJTT.
Each module is expected to discover different types of relationships among body keypoints from multiple individuals, as described in \cref{sec:method:2dto3d}. 
Particularly, IPST is expected to be beneficial for handling occlusions as it deals with every person all at once.

\begin{table}
  \resizebox{\linewidth}{!}{
    \setlength{\tabcolsep}{4pt}
    \centering
    \begin{tabular}{l|rrrr|rrrr}
    \toprule
     & \multicolumn{4}{c|}{MuPoTS-3D (full)} 
     & \multicolumn{4}{c}{MuPoTS-3D (heavy-occlusion)} \\
    Method &
    PCK$_\text{r}^\uparrow$ & PCK$_\text{a}^\uparrow$ & MPJVE$_\text{r}^\downarrow$ & MPJVE$_\text{a}^\downarrow$ &
    PCK$_\text{r}^\uparrow$ & PCK$_\text{a}^\uparrow$ & MPJVE$_\text{r}^\downarrow$ & MPJVE$_\text{a}^\downarrow$ \\
    \midrule
    w/o SPST
        & \textcolor{ForestGreen}{-47.3} & \textcolor{ForestGreen}{-40.1} & \textcolor{ForestGreen}{+3.5} & \textcolor{ForestGreen}{+23.0}
        & \textcolor{ForestGreen}{-45.9} & \textcolor{ForestGreen}{-39.6} & \textcolor{ForestGreen}{+2.4} & \textcolor{ForestGreen}{+8.2}
        \\
    w/o IPST 
        & \textcolor{ForestGreen}{-2.3} & \textcolor{ForestGreen}{-3.9} & \textcolor{ForestGreen}{+0.2} & \textcolor{ForestGreen}{+1.6}
        & \textcolor{ForestGreen}{-3.6} & \textcolor{ForestGreen}{-7.4} & \textcolor{ForestGreen}{+0.5} &
        \textcolor{ForestGreen}{+3.5}
        \\
    w/o SJTT 
        & \textcolor{ForestGreen}{-0.7} & \textcolor{ForestGreen}{-8.8} & \textcolor{ForestGreen}{+8.0} & \textcolor{ForestGreen}{+31.8}
        & \red{+0.2} & \textcolor{ForestGreen}{-4.3} & \textcolor{ForestGreen}{+5.7} & \textcolor{ForestGreen}{+23.4}
        \\
    \midrule
    Full  
        & 83.7 & 50.9 & 10.9 & 16.3
        & 82.1 & 47.2 & 12.2 & 17.3 \\

    \bottomrule
    \end{tabular}
} 
\caption{\centering Ablation on transformer blocks for 2D-to-3D lifting} 
\vspace{-0.1cm}
\label{tab:result_model_abl}
\end{table}

We conduct an ablation study for each Transformer, on the full and heavy-occlusion subset (2, 13, 14, 18, 20) of MuPoTS-3D.
From \cref{tab:result_model_abl}, we observe that 1) SPST has the biggest impact when removed, but we also see meaningful performance drops (\textcolor{ForestGreen}{marked green}) without IPST and SJTT; 2) Without IPST, PCK drops significantly, especially with heavy occlusion, indicating that it plays an important role in disambiguating multiple subjects; and 3) Without SJTT, MPJVE is significantly hurt, proving its role in learning temporal dynamics.
Overall, the results align with our expectations.

\noindent
\textbf{Augmentation Strategies.}
We further investigate the best data augmentation strategy proposed in \cref{sec:augmentation}.
We initially observe that augmenting only with PT+PR performs the best on the benchmark, leaving the impact of GPT and GPR marginal.
This might be caused by the limited diversity in the datasets. Most of cameras used for MuPoTS-3D are located near the ground, as the scenes are taken outside using a markerless MoCap system.
CMU-Panoptic has the same camera setting for both training and testing.
Thus, we conclude it is not a suitable setting to test generalizability of the augmentation method.

For this reason, we conduct this ablation study to evaluate zero-shot performance on a heterogeneous setting.
\cref{tab:ablation_mupots} compares on MuPoTS-3D the performance of our models that are trained on CMU-Panoptic, using multiple combinations of augmentation methods.
We observe that using more variety and larger scale of augmentations generally benefits.
A similar experiment is conducted on CMU-Panoptic, summarized in \cref{tab:ablation_panotic_2}.
Here, we further evaluate on videos of \textit{Haggling, Ultimatum} captured by different cameras (camera $6$, $13$) from the ones used for training (camera $16$, $30$).
(See \cref{fig:panoptic_camera_ablation} in Appendix \cref{sec:supp:quant} for illustration.)
Again, we confirm the benefit of the full augmentations and larger size.
These results prove that GPR benefits the model to be robust to camera view changes, aligning with our expectation. Without a limit, the proposed data augmentation may further improve the result with a larger training set.

\begin{table}
\resizebox{0.99\linewidth}{!}{ 
    \centering
    \begin{tabular}{cccc|c|cccc}
    \toprule
    PT & PR & GPT & GPR & Size & PCK$_\text{rel}^\uparrow$ & PCK$_\text{abs}^\uparrow$ & MPJVE$_\text{rel}^\downarrow$ & MPJVE$_\text{abs}^\downarrow$ \\  
    \midrule
    \checkmark & & &  & 0.5M & 50.8 & 13.6 & 13.7 & 35.8   \\
    \checkmark & \checkmark & &  & 0.5M & 58.9 & 24.9 & 12.7 & 22.4 \\
    \checkmark & \checkmark & \checkmark & & 0.5M & 54.9 & 22.0 & 12.7 & 27.2  \\
    \checkmark & \checkmark & \checkmark & \checkmark  & 0.5M &  \textbf{ 63.1} &\textbf{ 29.9 }& \textbf{11.8} & \textbf{21.1}  \\
    \midrule
    \checkmark & \checkmark & \checkmark & \checkmark  & 0.8M &  \textbf{ 67.1 }& \textbf{32.0} & \textbf{12.2} & \textbf{23.1}  \\
    \bottomrule
    \end{tabular}
} 
    \caption{\centering Ablation on augmentation strategies (MuPoTS-3D)}
    \label{tab:ablation_mupots}
    \vspace{-0.1cm}
\end{table}




\begin{table}
\resizebox{0.74\linewidth}{!}{ 
    \centering
    \begin{tabular}{cccc|c|cc}
    \toprule
    PT & PR & GPT & GPR &
    Size                                &
    MPJPE$_\text{rel}^\downarrow$  & MPJVE$_\text{rel}^\downarrow$\\
    \midrule
    \checkmark & & &
        & 0.5M
        & 142.9
        & \bf{3.1}\\
    \checkmark & \checkmark & &
        & 0.5M
        & 136.8
        & 6.1\\
    \checkmark & \checkmark & \checkmark &
        & 0.5M
        & 138.2
        & 5.3\\
    \checkmark & \checkmark & \checkmark & \checkmark
        & 0.5M
        & \bf{67.2}
        & 3.7\\
    \midrule
    \checkmark & \checkmark & \checkmark & \checkmark
        & 0.8M
        & \bf{58.4}
        & \bf{2.7} \\
    \bottomrule
    \end{tabular}
} 
\caption{
    Ablation on camera setting (CMU-Panoptic)
} 
\label{tab:ablation_panotic_2}
\vspace{-0.1cm}
\end{table}


\begin{figure*}
  \centering{\includegraphics[width=\textwidth]{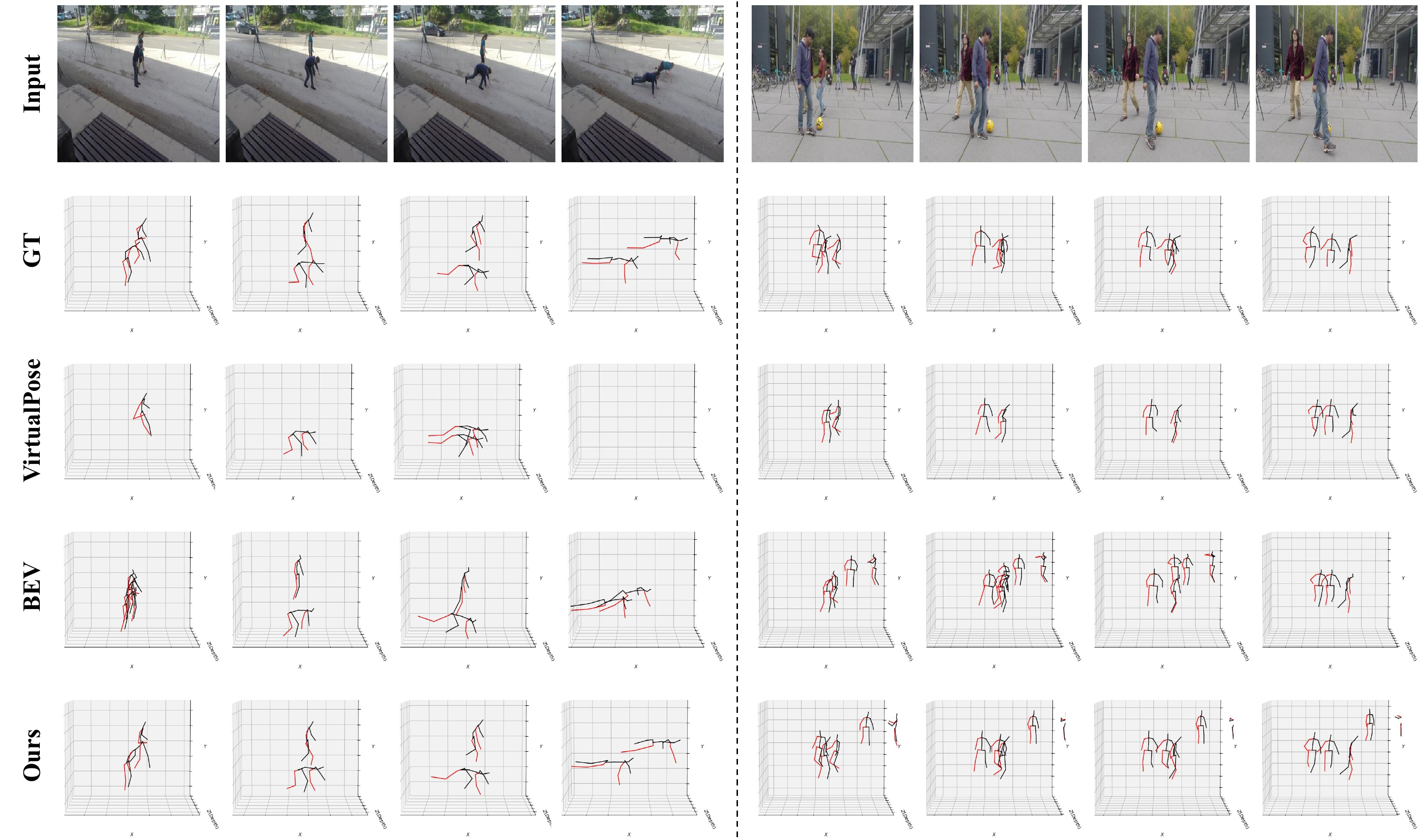}}
  \caption{\textbf{Qualitative Results on MuPoTS-3D} of ours and recent baselines. The challenges are \textit{(Left)} unusual camera distance, and \textit{(Right)} heavy occlusions. Ours works most successfully despite the challenges. \vspace{-0.2cm}
  }
  \label{fig:mupots_qual}
\end{figure*}

\begin{figure}
  \centering
  \includegraphics[width=0.99\linewidth]{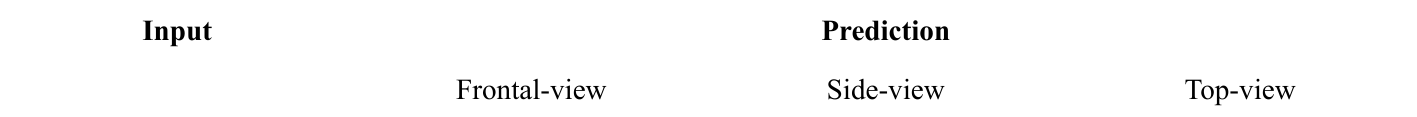}
  \includegraphics[width=0.99\linewidth]{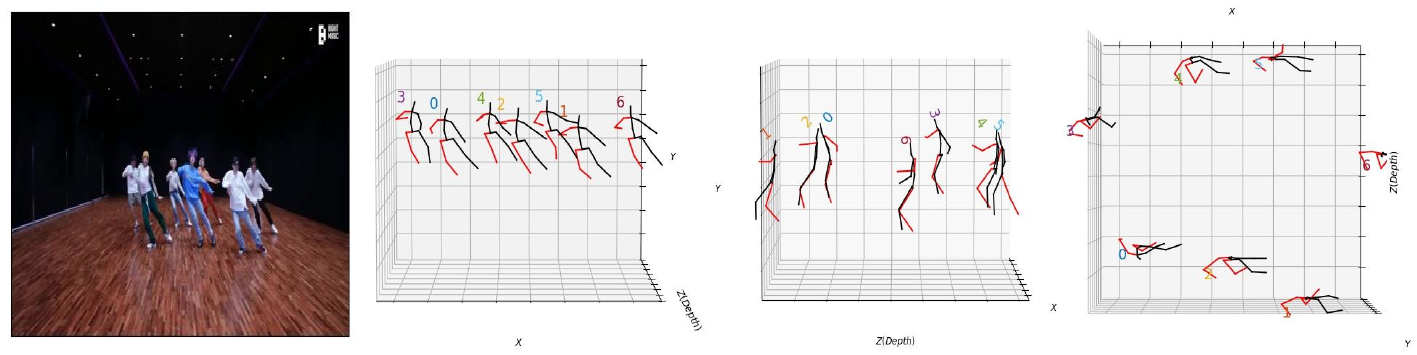}
  \includegraphics[width=0.99\linewidth]{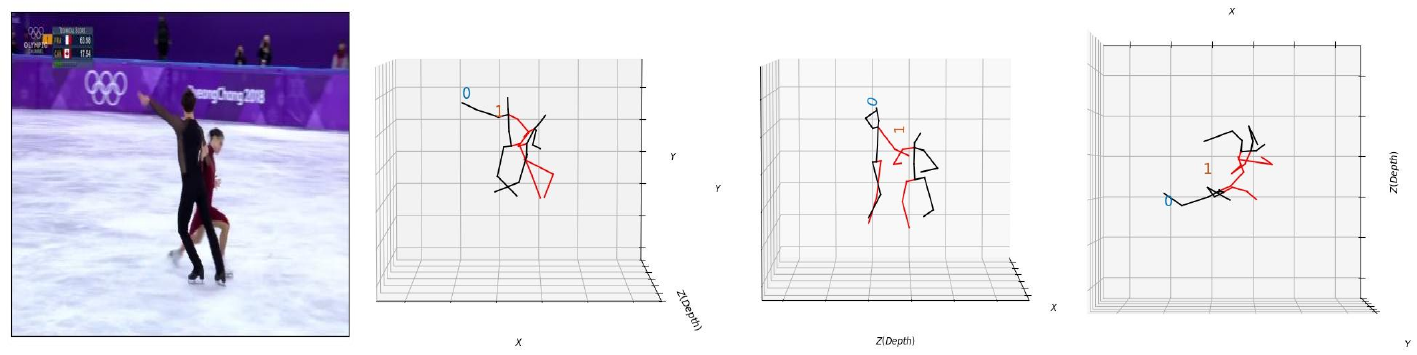}
  \includegraphics[width=0.99\linewidth]{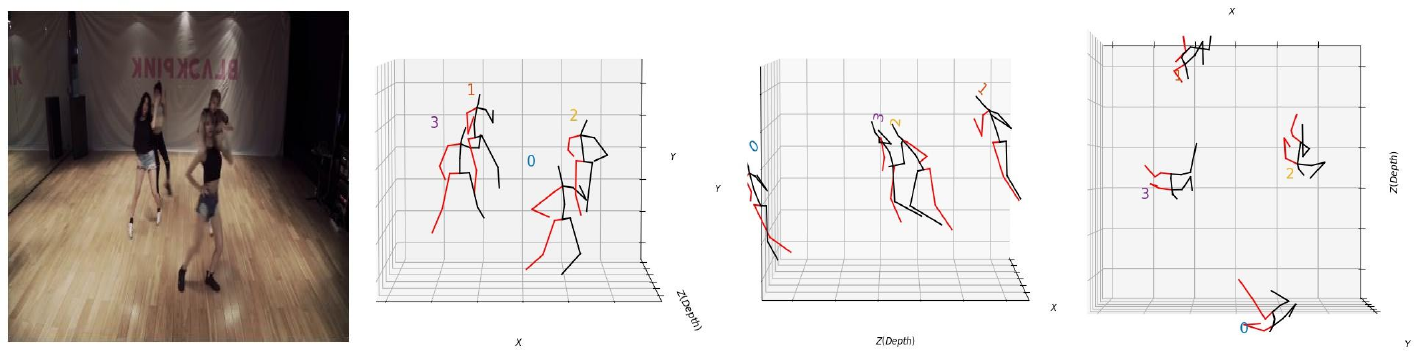}
  
  \caption{\centering Demonstration of \propose\ on in-the-wild videos. \vspace{-0.2cm}}
  \label{fig:in-the-wild}
\end{figure}


\begin{table}
\resizebox{0.82\linewidth}{!}{ 
    \centering
    \begin{tabular}{cc|cccc}
    \toprule
    
     Sim. & Thresh.
     & PCK$_\text{rel}^\uparrow$ & PCK$_\text{abs}^\uparrow$ & MPJVE$_\text{rel}^\downarrow$ & MPJVE$_\text{abs}^\downarrow$ \\ 
    \midrule
    
      &  & 77.0 & 25.1 & 27.9 & 59.4 \\
    \checkmark &  & \textbf{83.7} & 48.7 & 12.0 & 18.8 \\
    \checkmark & \checkmark & \textbf{83.7} & \textbf{50.9} & \textbf{10.9} & \textbf{16.3} \\
    
    \bottomrule
    \end{tabular}
} 
    \caption{Ablation on occlusion handling (MuPoTS-3D)} 
    \label{tab:ablation_mupots_aug_abl}
    \vspace{-0.1cm}
\end{table}

\vspace{0.1cm} \noindent
\textbf{Occlusion Handling.}
We investigate the effect of our occlusion handling strategy: 1) simulating occlusion by adding noise to or dropping out occluded keypoints during augmentation with the person volumetric model (Sim.), and 2) filtering out 2D pose prediction with low confidence at inference (Thresh.). 
As shown in \cref{tab:ablation_mupots_aug_abl}, simulating occlusion during augmentation significantly improves the overall performance, especially for absolute position prediction.
This indicates that training the model not to be confused by noisy information at training is actually effective.
Thresholding at inference gives further incremental improvement in most metrics.


\subsection{Qualitative Results} 
\label{sec:exp_qual}

\cref{fig:mupots_qual} shows the estimated 3D poses from ours and several baselines on two challenging sequences of MuPoTS-3D.
Baseslines fail to estimate exact depth when the camera distance is unusual (\textit{Left}), and totally miss individuals when heavy occlusion occurs (\textit{Left, Right}).
However, \propose\ successfully reconstructs every individual in the 3D space, even including omitted persons in GT annotations far away from the camera.

Beyond the benchmark datasets, we evaluate \propose\ on a lot more challenging in-the-wild scenarios, \eg, group dancing video or figure skating.
\cref{fig:in-the-wild} demonstrates the performance of our model on a few examples of in-the-wild videos. 
In spite of occlusions, we see that \propose\ precisely estimates poses of multiple people.
To the best of our knowledge, this is the first work to present such accurate and smooth 3DMPPE results on in-the-wild videos.

Meanwhile, ours is somewhat sensitive to the performance of the 2D pose tracker.
The tracker we use often fails when extreme occlusion occurs or persons look too homogeneous.
More examples including some failure cases are provided in \cref{sec:supp:wild}.




\vspace{-0.1cm}

\section{Summary}
\label{sec:summary}

\vspace{-0.2cm}

In this paper, we propose \propose, the first realization of a \textit{seq2seq} 2D-to-3D lifting model for 3DMPPE,  
powered by geometry-aware data augmentations.
Considering important geometric factors like the ground plane orientation and occlusions, our proposed augmentation scheme benefits the model to be robust on a variety of views, overcoming the long standing data scarcity issue.
The effectiveness of our approach is verified not only by achieving state-of-the-art performance on public benchmarks, but also by demonstrating more natural and smoother results on various in-the-wild videos.
We leave a few interesting extensions, size disambiguation or more proactive augmentation on the pose itself, as a future work.



\vspace{0.1cm}
{ \small
\noindent\textbf{Acknowledgement.} This work was supported by the New Faculty Startup Fund from Seoul National University and by National Research Foundation (NRF) grant (No. 2021H1D3A2A03038607/50\%, 2022R1C1C1010627/20\%, RS-2023-00222663/10\%) and Institute of Information \& communications Technology Planning \& Evaluation (IITP) grant (No. 2022-0-00264/20\%) funded by the government of Korea.}

{\small
\bibliographystyle{ieee_fullname}
\bibliography{main}
}

\clearpage
\appendix

\pagenumbering{roman}
\renewcommand\thetable{\Roman{table}}
\renewcommand\thefigure{\Roman{figure}}
\setcounter{table}{0}
\setcounter{figure}{0}

\section*{Appendix}

\section{Off-the-shelf Models}
\label{sec:supp:details}


\subsection{2D Pose Estimation}
As mentioned in \cref{sec:exp:setting}, we adopt a pretrained HRNet~\cite{sun2019deep} model for 2D pose estimation.
Since the model was originally trained on MSCOCO~\cite{mscoco}, which uses slightly different format of human body keypoints, we fine-tune the model to convert the keypoints to the desired format. (Note that MSCOCO, MuPoTS-3D, and CMU-Panoptic use different keypoints, while MuPoTS-3D, MuCo-3DHP, and MPI-INF-3DHP share the same one.)
The model is fine-tuned on the MuCo-3DHP dataset for MuPoTS-3D testing, and fine-tuned on CMU-Panoptic training dataset for CMU-Panoptic testing, respectively.
Each of which is fine-tuned for 20 epochs with a learning late of $10^{-4}$, which is 10 times smaller compared to the original learning rate at training.

\subsection{2D Pose Tracking}
In \cref{sec:exp:setting}, we mention that we use ByteTrack~\cite{zhang2022bytetrack} for re-identification of each individual.
We also merge the appearance gallery idea \cite{Wojke2018deep} to consider appearance variation caused by movements. While tracking individuals frame by frame, the most recent 100 appearance features are stored in their tracklet.
For measuring similarities, in total three similarities are used. In addition to the appearance feature and IoU, we also consider the similarity between poses as well.

We observe clean and accurate tracking is important for end-to-end performance.
Specifically, performance of our model is quite sensitive to the tracking result, especially, to the weights among 3 similarities above.
Giving a higher weight to the appearance similarity might help the model to match re-appearing individuals after a period of heavy-occlusion, meanwhile it does not consider about their location. 
In contrast, giving a higher weight to IoU or pose similarity might help the model to accurately track the motion dynamics of individuals, but as a trade-off, it confuses the model to match re-appearing individuals.
We empirically find the optimal weights and use \{0.4, 0.3, 0.3\} for each appearance, IoU, and pose similarity, respectively.


One thing to note is that the end-to-end evaluation metrics are not significantly affected by a few tracking failures, as it usually match each predicted individual with the ground truth frame-by-frame.
For a real-world application, largely disturbing tracklets might be omitted and only some clean tracklets could be chosen for the sake of reliability.

\section{Details on Data Augmentation}
\label{sec:supp:augmentation}

\joonseok{The data augmentation hyperparameters, $\alpha, \beta, \gamma, \theta, \varphi$, are empirically chosen, considering typical movement range of a person for translations and approximate angular distance between cameras of the source dataset for rotations, respectively.}
$\alpha, \beta$ are sampled from a Gaussian $\mathcal{N}(0, 6.0^2)$ and $\gamma$ is randomly chosen among $\{-1.0, 0, 1.5, 3.0\}$, where the unit is meters. $\theta$ is uniformly sampled within $[-\pi/4, \pi/4]$, and $\varphi$ is randomly chosen among $\{-\pi/6, 0, \pi/6\}$.

\vspace{0.1cm}
\textbf{Validity of a Training Example.}
Once a training example is generated, we check its validity.
First of all, the depth $z$ of all target 3D keypoints should be positive.
Otherwise, a subject with negative depth will 
appear flipped both vertically and horizontally after projection.

Also, as we constrain the number of people appearing in the scene to be consistent throughout the temporal receptive field (\textit{i.e.} people are assumed not to be jumping in or fading out), we force the resulting trajectory to be entirely located within the 2D frames.
Precisely, we keep the root key points to appear within the image boundary but let other joints potentially be out of the scene.
For this, we might naively filter out examples that violate the constraints and regenerate, but this is not efficient.
Instead, we apply PR, GPT, and GPR first, and PT at the last. Unlike other operations, we can constrain the feasible range for PT individually, satisfied simply by solving a constrained linear programming:
\begin{align}
    0 &\leq f_u \frac{x + \Delta x}{z + \Delta z} + c_u < W,\nonumber \\
    \hspace{0.2cm}
    0 &\leq f_v \frac{y + \Delta y}{z + \Delta z} + c_v < H, \\
    \hspace{0.2cm}
    0 &\leq z + \Delta z,\nonumber
    \label{eq:pt_constraints}
\end{align}
where $(x, y, z)$ is an original root joint in the 3D space, $(\Delta x, \Delta y, \Delta z)$ is the amount of displacement applied to this subject, converted from $(\alpha,  \beta)$ on the basis $\{\mathbf{b}_1, \mathbf{b}_2\}$ to the standard basis ($\mathbf{e}_1, \mathbf{e}_2, \mathbf{e}_3$), and $W, H$ is the width and height of the image.

\section{More Quantitative Results}
\label{sec:supp:quant}

\begin{table*}
\resizebox{\linewidth}{!}{ 
    \centering
    \begin{tabular}{l|c|cccccccccccccccccccc}
    \toprule
    Method  & PCK$_\text{rel}$(\%)$\uparrow$ & TS1 &TS2 &TS3 &TS4 &TS5 &TS6 &TS7 &TS8 &TS9 &TS10 &TS11 &TS12 &TS13 &TS14 &TS15 &TS16 &TS17 &TS18 &TS19 &TS20  \\ 
    \midrule

    
    SingleStage~\cite{Jin_2022_CVPR}    & 80.9   & -- & -- & -- & -- & -- & -- & -- & -- & -- & -- & -- & -- & -- & -- & -- & -- & -- & -- & -- & -- \\
    
    SMAP~\cite{zhen2020smap}                                & 73.5  & 88.8 & 71.2 & 77.4 & 77.7 & 80.6 & 49.9 & 86.6 & 51.3 & 70.3 & 89.2 & 72.3 & 81.7 & 63.6 & 44.8 & 79.7 & 86.9 & 81.0 & 75.2 & 73.6 & 67.2 \\
    
    SDMPPE~\cite{Moon_2019_ICCV_3DMPPE}                              & {81.8} & \bf{94.4} & 77.5 & 79.0 & 81.9 & 85.3 & 72.8 & 81.9 & \bf{75.7} & \bf{90.2} & \textbf{90.4} & 79.2 & 79.9 & 75.1 & 72.7 & 81.1 & 89.9 & \bf{89.6} & 81.8 & 81.7 & 76.2 \\
    
    \propose\ (Ours)          & \bf{83.7} & 92.0 & \bf{80.2} & \bf{83.7} & \bf{84.0} & \textbf{85.4} & \bf{75.1} & \bf{91.5} & 74.3 & 70.7 & 88.4 & \bf{85.6} & \bf{86.5} & \bf{83.1} & \bf{77.1} & \bf{82.8} & \textbf{90.8} & 86.8 & \bf{87.5} & \bf{85.7} & \bf{82.6}  \\
    
    \toprule
    Method  & PCK$_\text{abs}$(\%)$\uparrow$ & TS1 &TS2 &TS3 &TS4 &TS5 &TS6 &TS7 &TS8 &TS9 &TS10 &TS11 &TS12 &TS13 &TS14 &TS15 &TS16 &TS17 &TS18 &TS19 &TS20  \\ 
    \midrule
    VirtualPose~\cite{VirtualPose2022}  & 44.0 & -- & -- & -- & -- & -- & -- & -- & -- & -- & -- & -- & -- & -- & -- & -- & -- & -- & -- & -- & -- \\
    
    SingleStage~\cite{Jin_2022_CVPR}    & 39.3   & -- & -- & -- & -- & -- & -- & -- & -- & -- & -- & -- & -- & -- & -- & -- & -- & -- & -- & -- & -- \\
    
    SMAP~\cite{zhen2020smap}                                & 35.2	& 21.4	& 22.7	&  58.3	& 27.5	& 37.3	& 12.2	& 49.2	& 40.8	& \bf{53.1}	& 43.9	& \textbf{43.2}	& 43.6	& 39.7	& 28.3	& 49.5	& 23.8	& 18.0	& 26.9	& 25.0	& 38.8 \\
    
    SDMPPE~\cite{Moon_2019_ICCV_3DMPPE}                              & 31.5	& \bf{59.5}	& \bf{44.7}	& 51.4	& 46.0	& 52.2	& 27.4	& 23.7	& 26.4	& 39.1	& 23.6	& 18.3	& 14.9	& 38.2	& 26.5	& 36.8	& 23.4	& 14.4	& 19.7	& 18.8	& 25.1 \\
  
    \propose\ (Ours)          & \bf{50.9}	& 50.1	& 42.1	& \textbf{71.0}	& \bf{60.5}	& \bf{58.6}	& \bf{50.4}	& \textbf{66.9}	& \textbf{41.5}	& 50.0	& \bf{69.6}	& 42.3	& \textbf{49.2} & \bf{63.2}	& \bf{49.3}	& \bf{69.0}	& \bf{35.6}	& \bf{36.9}	& \bf{35.3}	& \bf{29.3}	& \textbf{46.3}  \\

    \bottomrule
    \end{tabular}
} 
    \caption{
        \textbf{Quantitative Comparison on MuPoTS-3D for Individual Test Videos.}
        The best scores are marked in boldface.
    } 
    \label{tab:result_ts}
\end{table*}

\cref{tab:result_ts} lists the performance of our model and baselines on individual test videos in MuPoTS-3D.
We observe that our proposed method, \propose, outperforms baselines on most videos, especially when severe occlusion occurs (TS 2, 13, 14, 18, 20).

\begin{figure}
  \centering
  \includegraphics[width=1.0\linewidth]{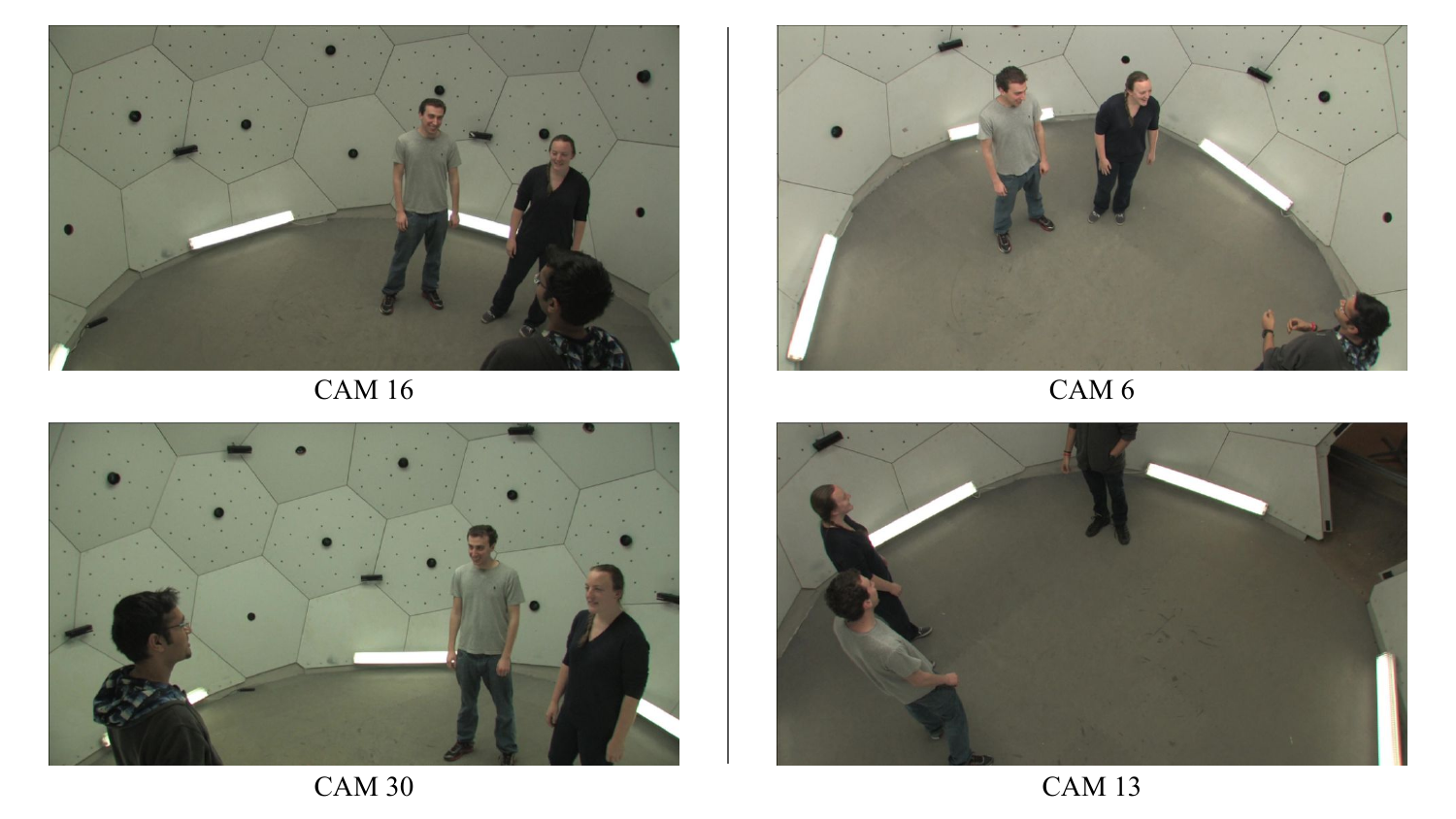}
  \caption{\textbf{Camera view points of CMU-Panoptic.} (\textit{Left}) Cameras used in conventional benchmark for both training and testing. (\textit{Right}) Cameras used for our experiment in \cref{sec:exp:ablation}.}
  \label{fig:panoptic_camera_ablation}
\end{figure}

For reference, \cref{fig:panoptic_camera_ablation} illustrates the conventional camera settings in CMU-Panoptic, and the ones we use in \cref{sec:exp:ablation}.

\section{More In-the-wild Examples}
\label{sec:supp:wild}

\begin{figure*}
  \centering
  \includegraphics[width=0.98\textwidth]{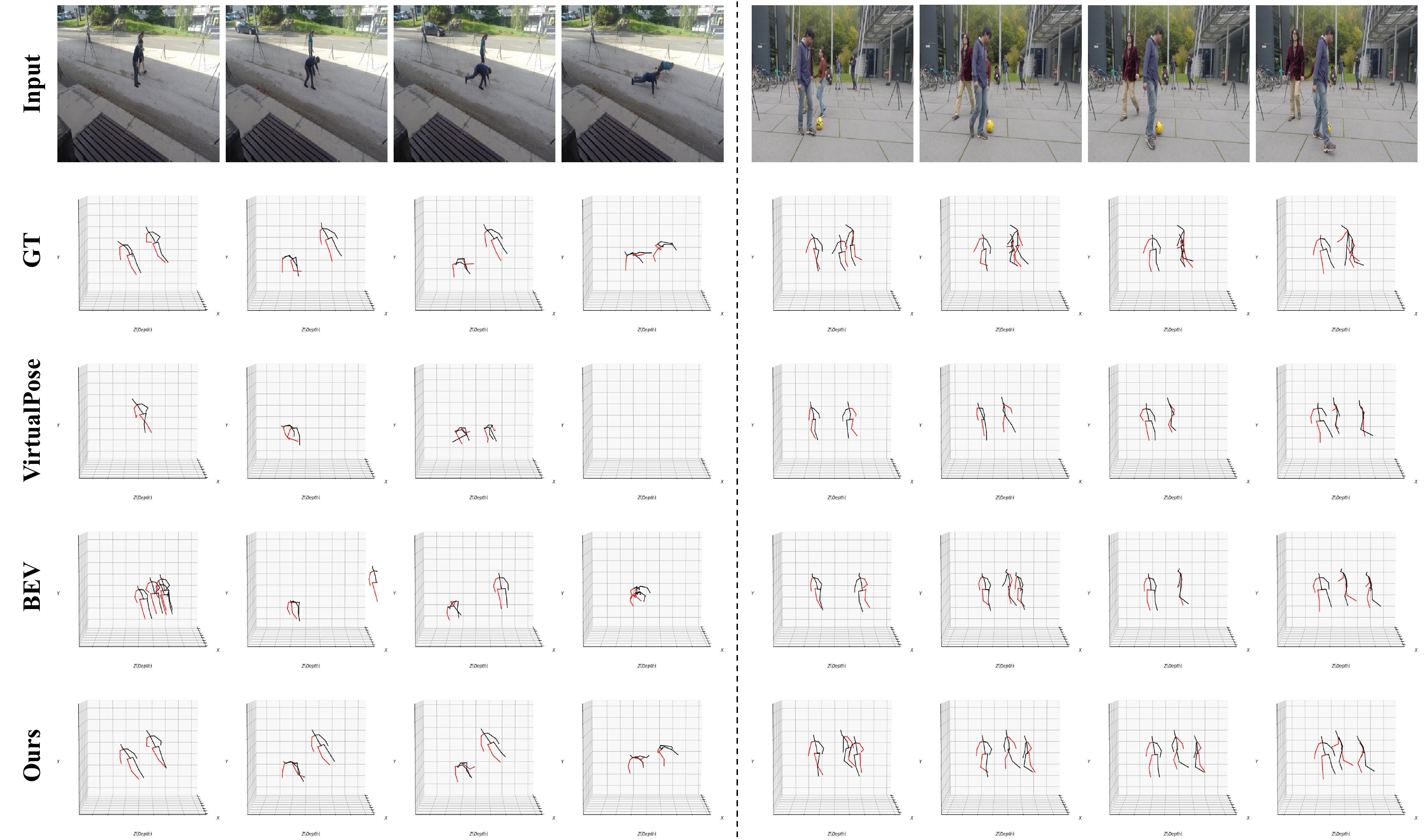}\\
  \caption{\textbf{Qualitative Results on MuPoTS-3D} of ours and recent baselines \textbf{(Side View)}.}
  \label{fig:mupots_qual_side}
\end{figure*}

\begin{figure*}
  \centering
  \includegraphics[width=0.98\textwidth]{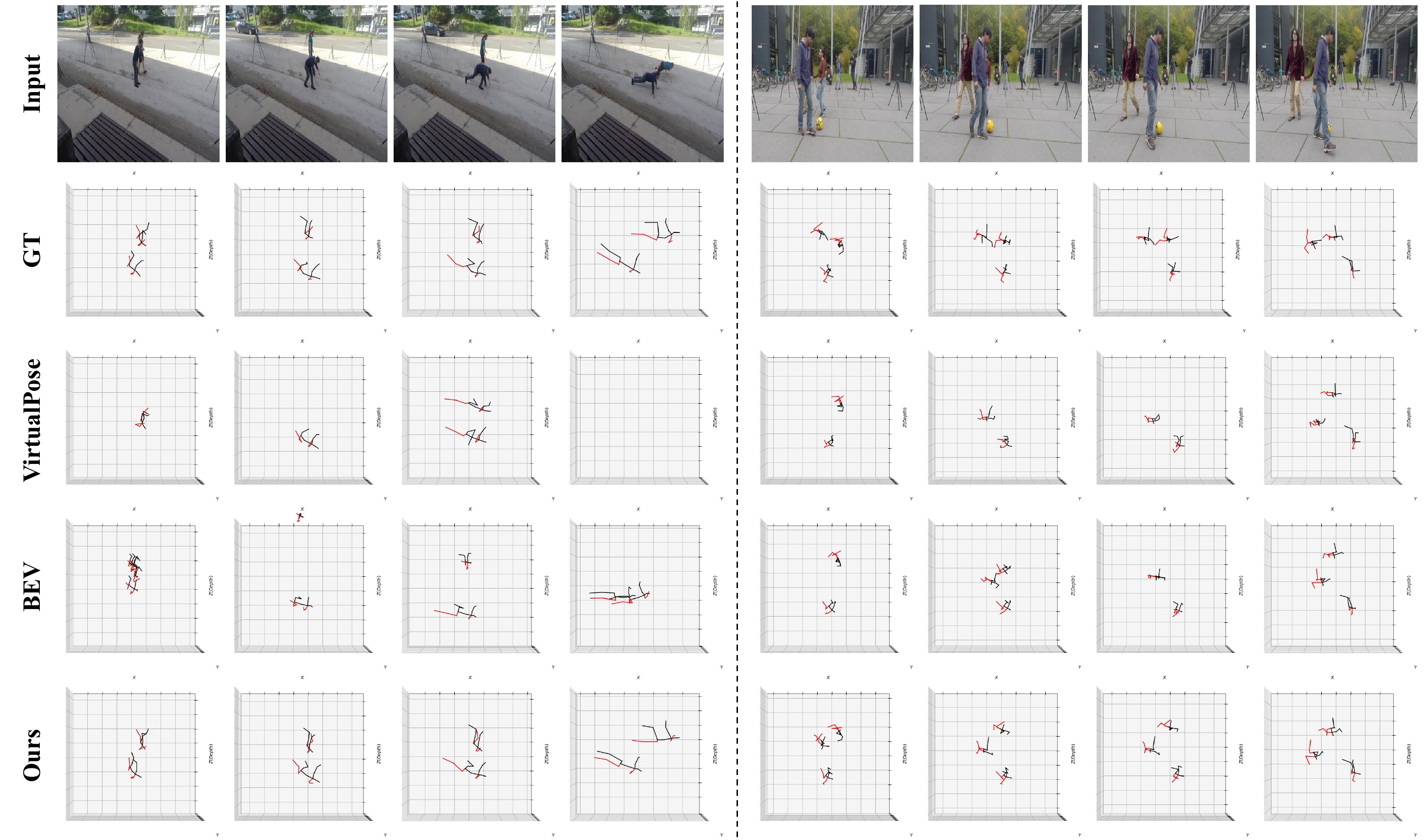}\\
  \caption{\textbf{Qualitative Results on MuPoTS-3D} of ours and recent baselines \textbf{(Top View)}.}
  \label{fig:mupots_qual_top}
\end{figure*}

\cref{fig:mupots_qual_side}--\ref{fig:mupots_qual_top} illustrate qualitative results on MuPoTS-3D of ours and baselines from side view, and top view. They appeal that the depth estimation of \propose\ is more accurate and smooth.

\begin{figure*}
  \centering
  \includegraphics[width=0.98\textwidth]{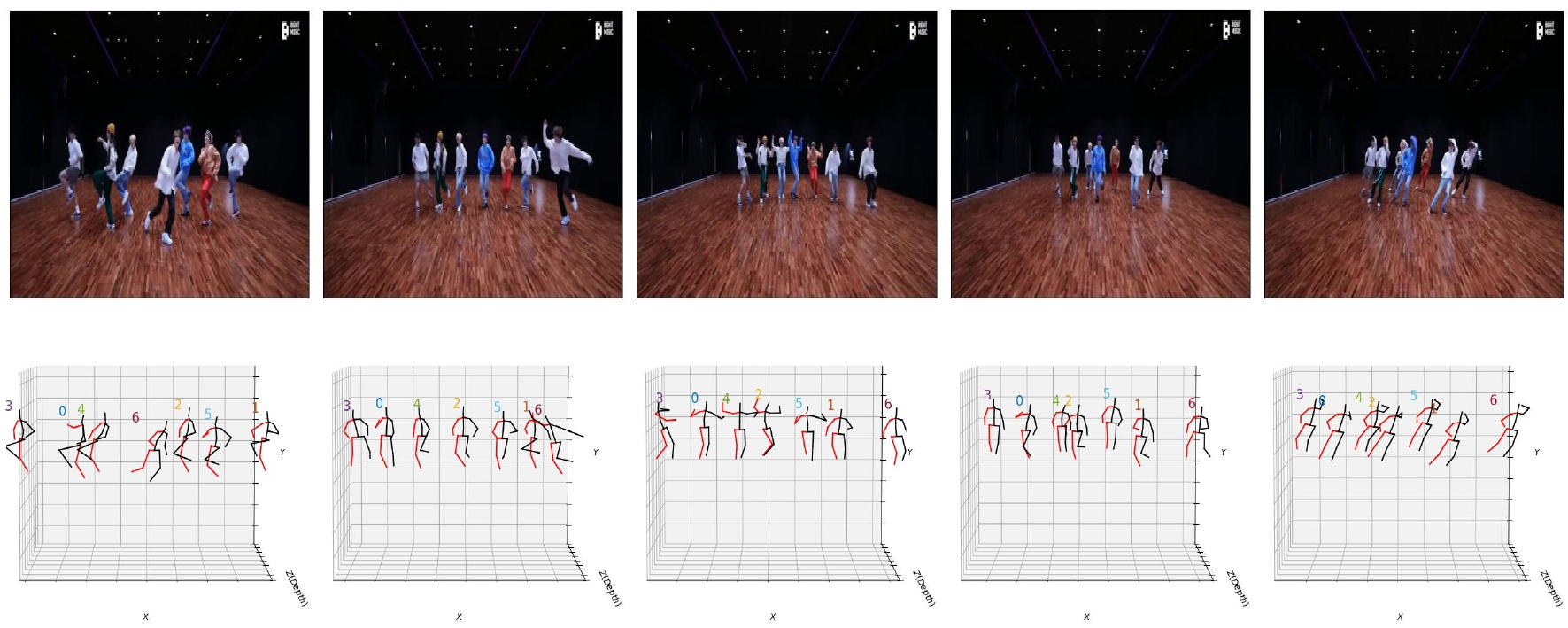}\\
  \includegraphics[width=0.98\textwidth]{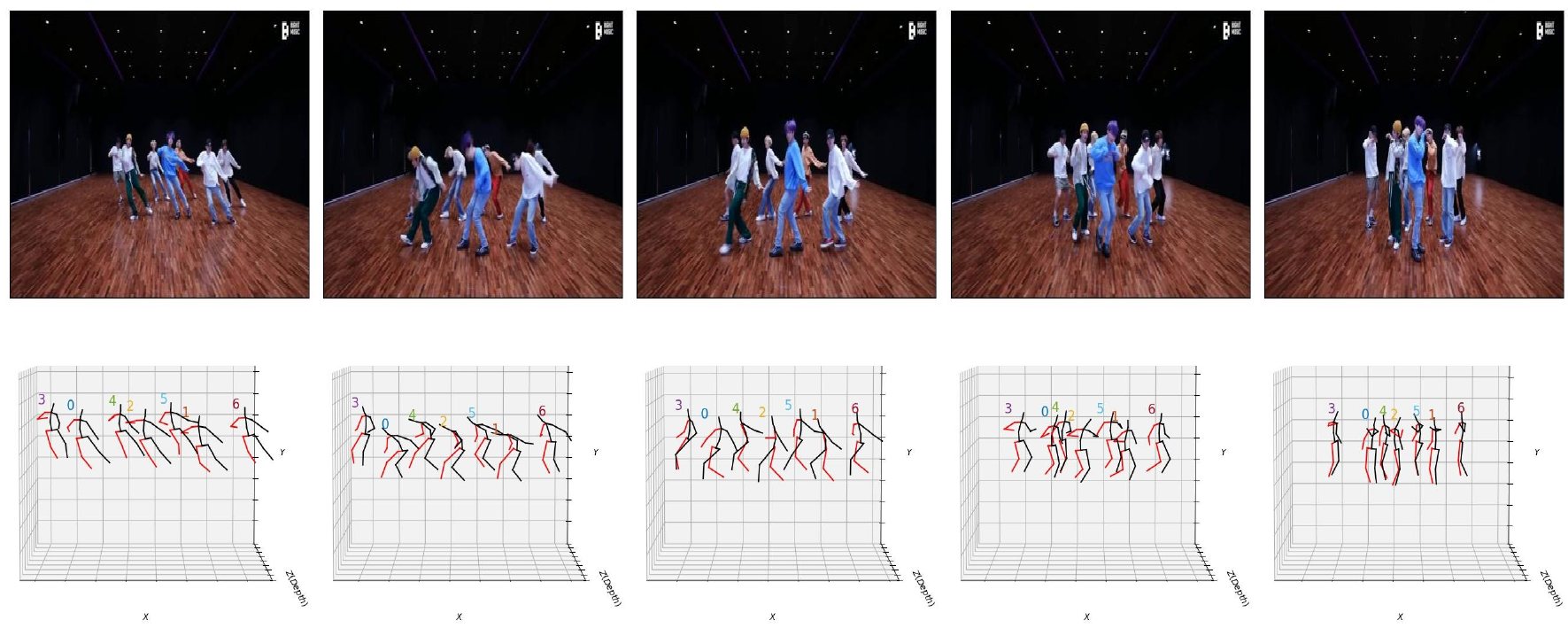}\\
  \caption{Additional Examples of in-the-wild inference (1/8) -- Group dance \textbf{(Massive movements and occlusions)}}
  \label{fig:additional_examples_first}
\end{figure*}

\begin{figure*} 
  \centering
  \includegraphics[width=0.98\textwidth]{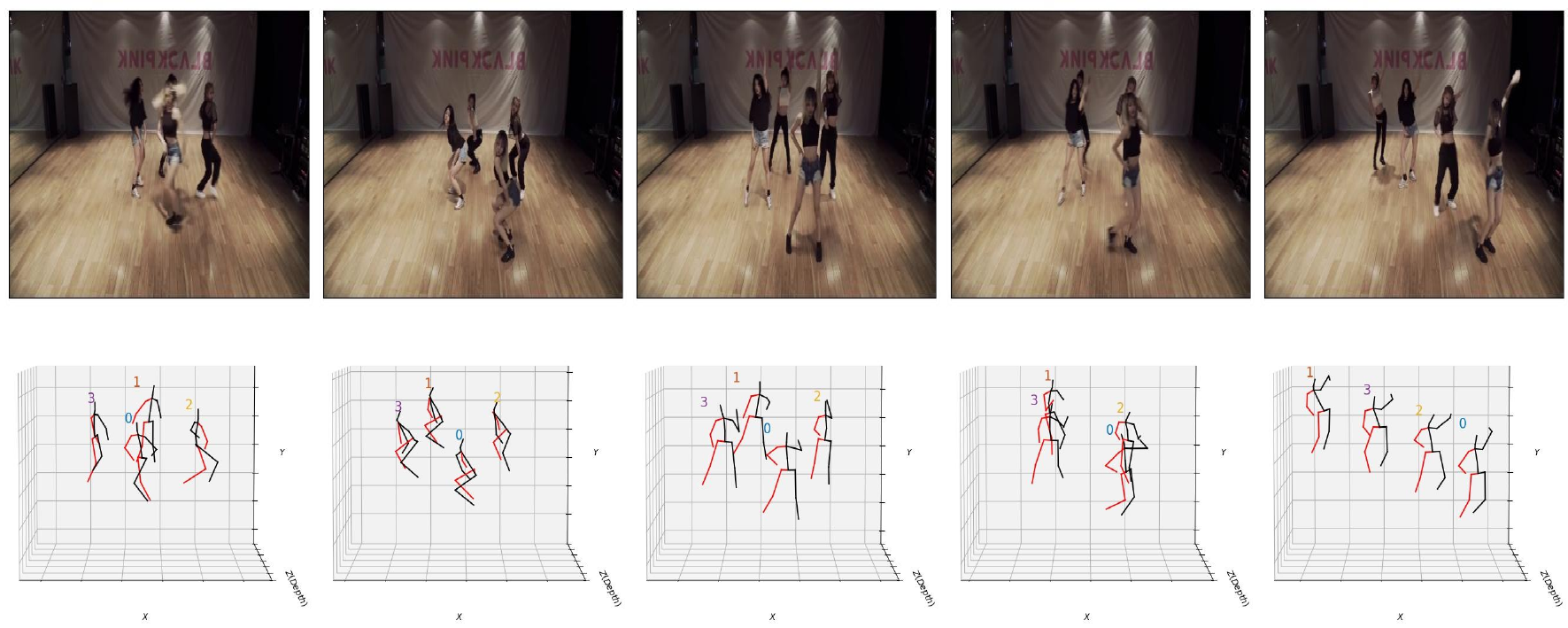}
  \includegraphics[width=0.98\textwidth]{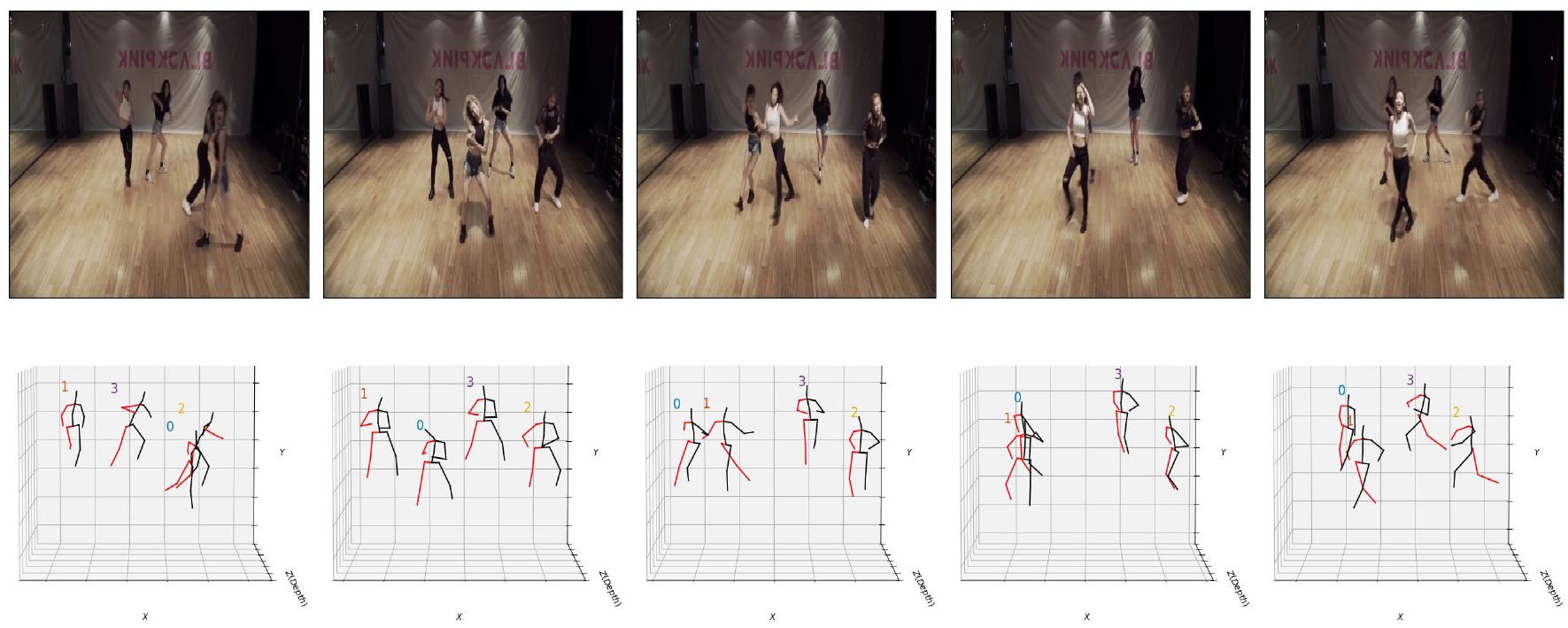}
  \caption{\centering Additional Examples of in-the-wild inference (8/8) -- Girl group dance practicing \textbf{(Massive movements and occlusions)}}
  \label{fig:additional_examples_last}
 \end{figure*}

\begin{figure*}
  \centering
  \includegraphics[width=0.98\textwidth]{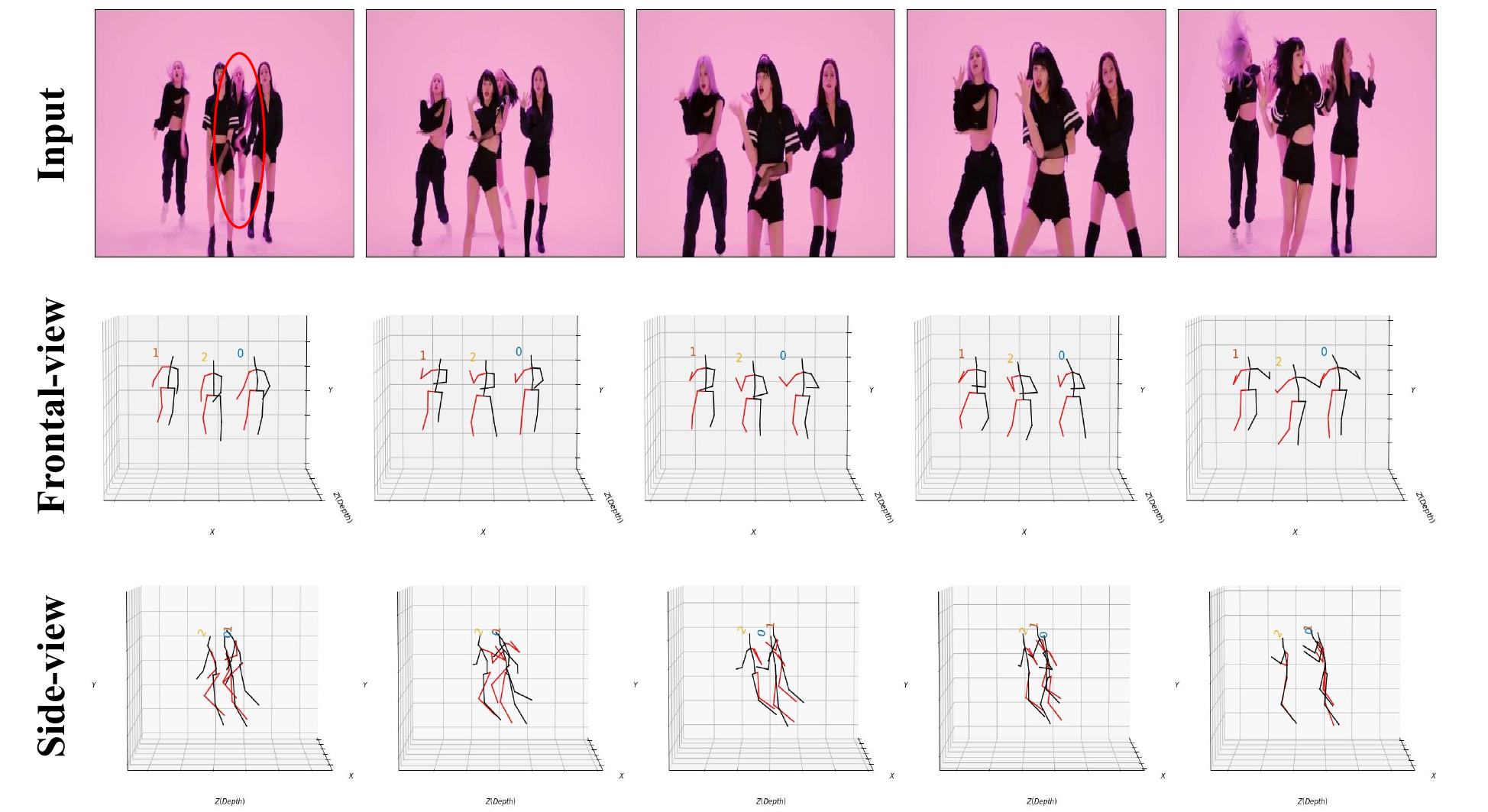} \\
  \vspace{0.7cm}
  \includegraphics[width=0.98\textwidth]{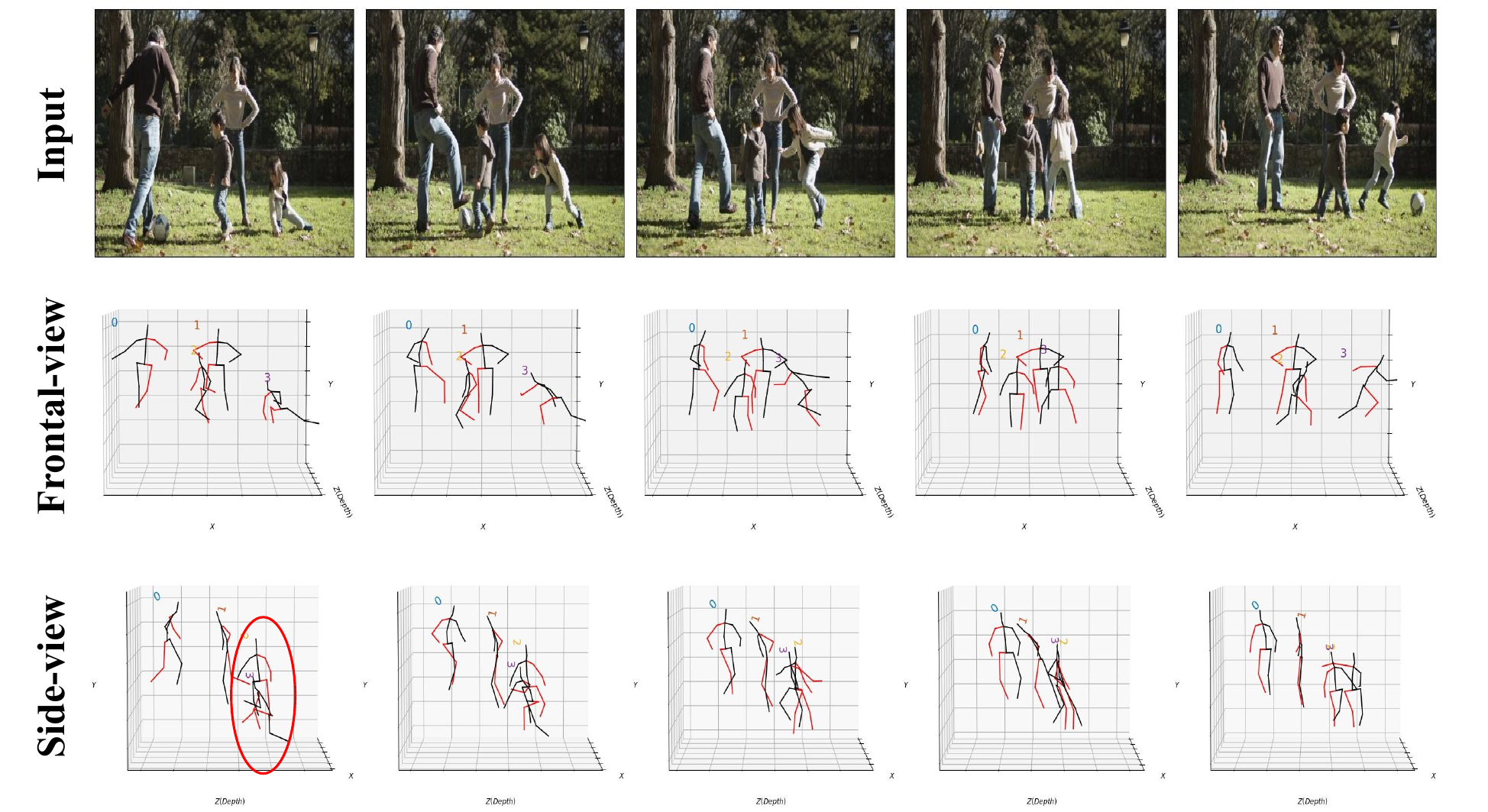} \\
  \caption{\textbf{Failure Cases}. \textbf{(Top)} Tracking Failure, where one person is not totally tracked due to heavy occlusion. \textbf{(Bottom)} Depth Ambiguity, where the depth for children is wrongly estimated which can be checked in the side view.}
  \label{fig:additional_examples_failure}
\end{figure*}

\cref{fig:additional_examples_first}--\ref{fig:additional_examples_last} illustrate more qualitative results of our model on several challenging in-the-wild videos.
We present the results of 10 frames from the frontal view per video to demonstrate both accuracy and smoothness.
It robustly operates even in highly challenging situations, such as heavy occlusions, dynamic motions, and non-static camera movement.
The results from other views are provided in the demo video at \href{https://www.youtube.com/@potr3d}{https://www.youtube.com/@potr3d}.
To create this video, we use a \propose\ model trained on the augmented dataset from MPI-INF-3DHP with Aug4, and assume a general focal length (\textit{i.e.}, 1500) to denormalize the depths.
At last, some examples in \cref{fig:additional_examples_failure} include additional failure cases, caused by tracking failure, and depth ambiguity.


\clearpage

\begin{figure*} 
  \centering
  \includegraphics[width=0.98\textwidth]{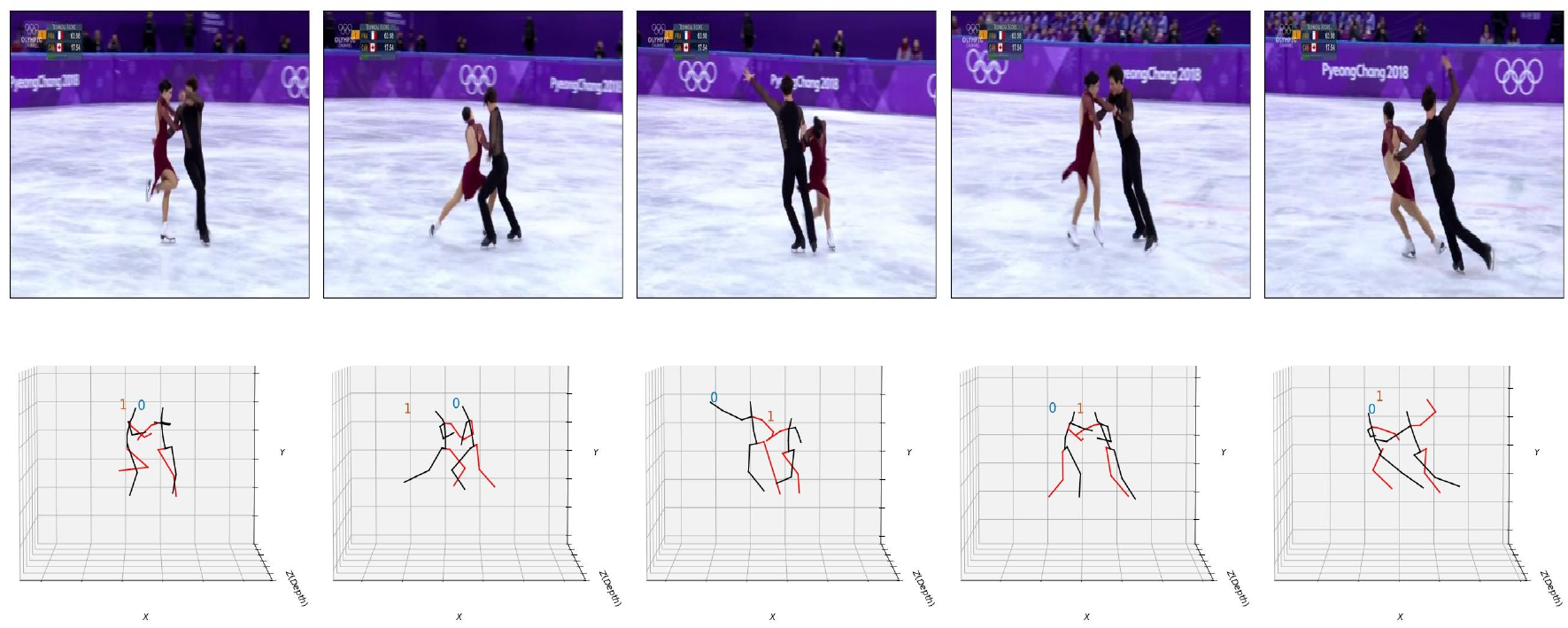}\\
  \includegraphics[width=0.98\textwidth]{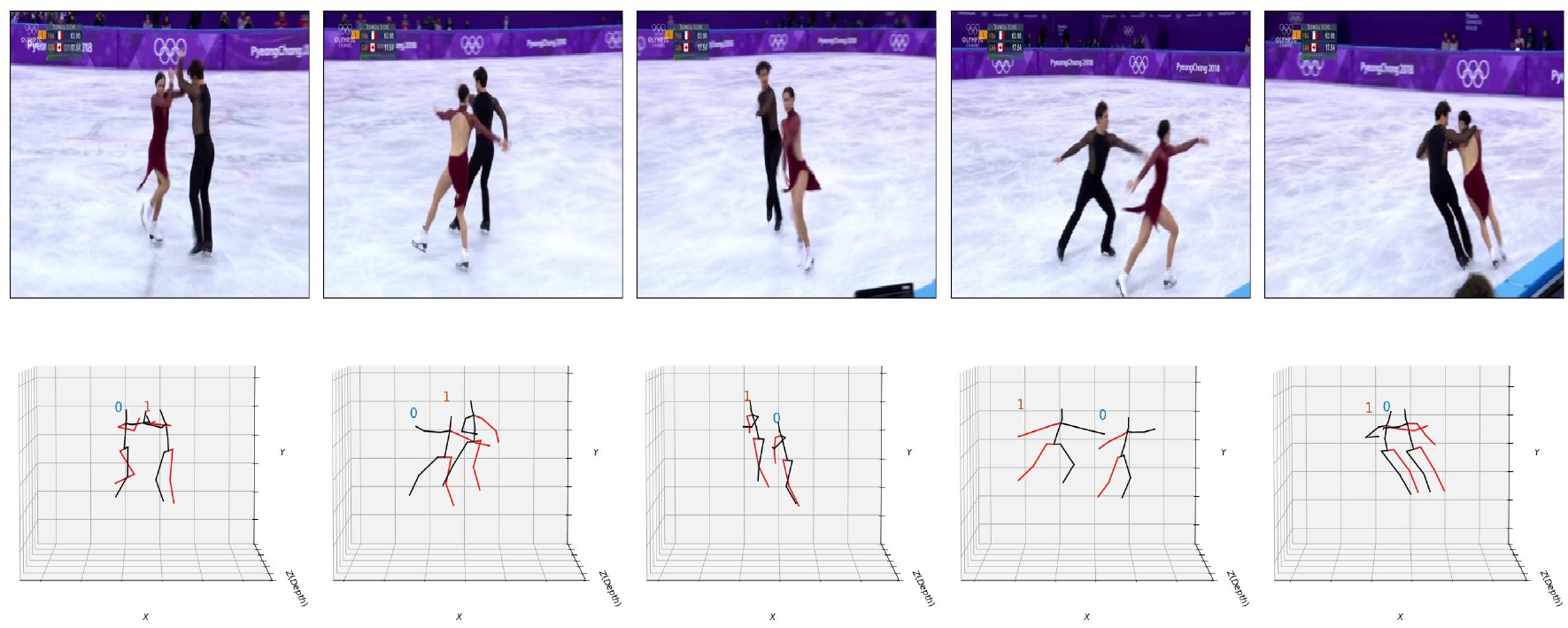}\\
  \caption{Additional Examples of in-the-wild inference (2/8) -- Figure skating \textbf{(Rampant movements and occlusions)}}
\end{figure*}

\clearpage

\begin{figure*} 
  \centering
  \includegraphics[width=0.98\textwidth]{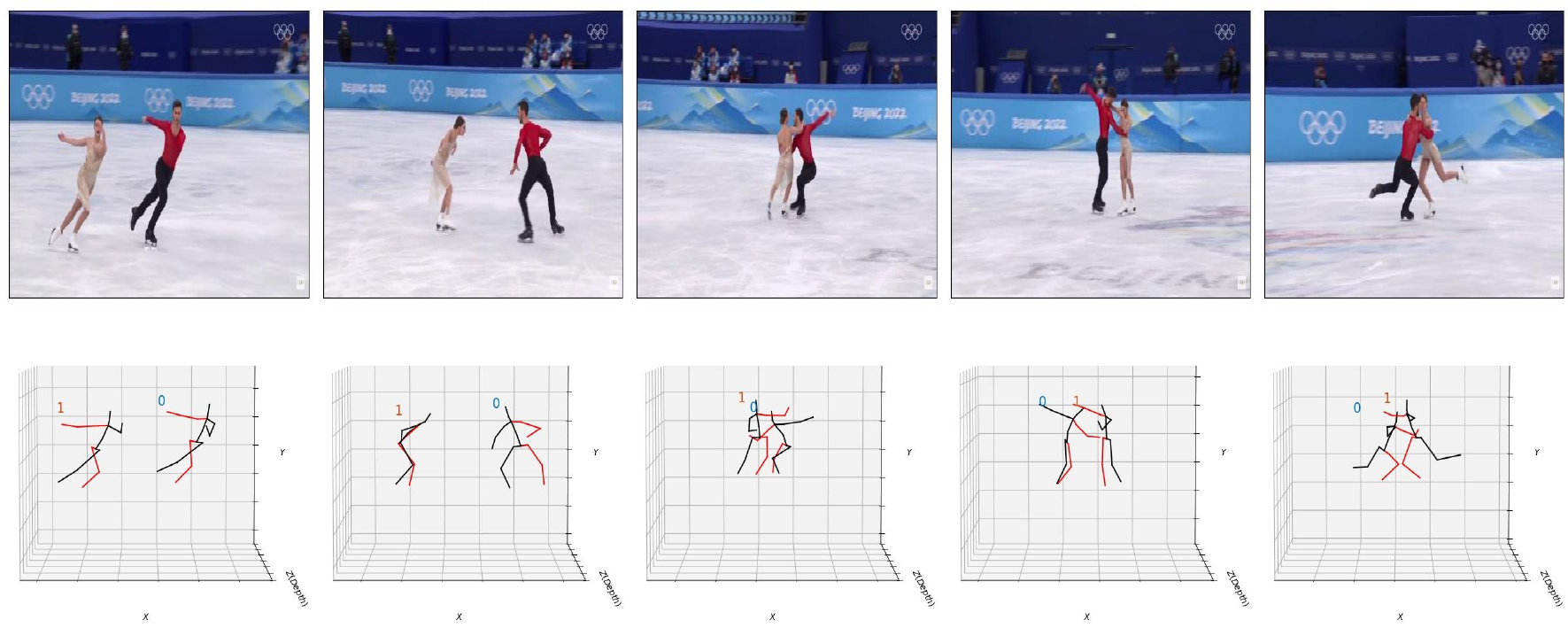}\\
  \includegraphics[width=0.98\textwidth]{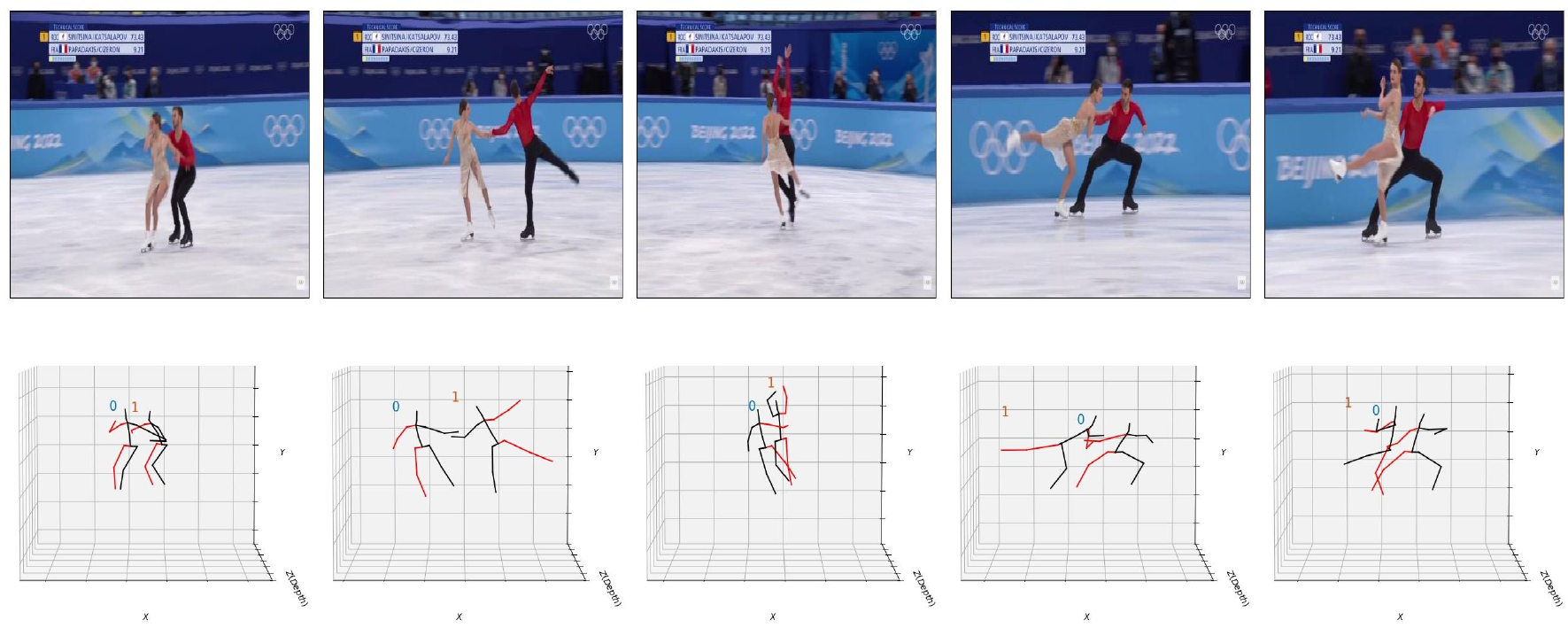}\\
  \caption{Additional Examples of in-the-wild inference (3/8) -- Figure skating \textbf{(Rampant movements and occlusions)}}
\end{figure*}

\clearpage

\begin{figure*} 
  \centering
  \includegraphics[width=0.98\textwidth]{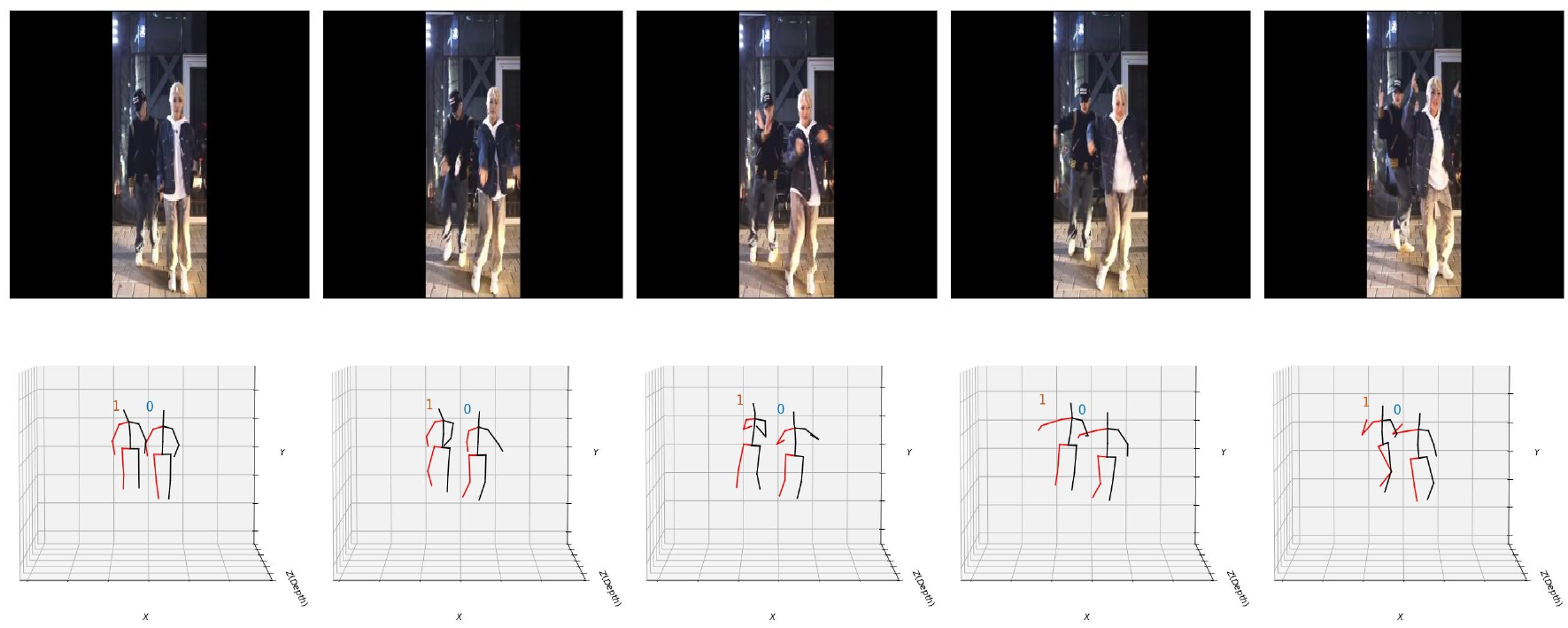}\\
  \includegraphics[width=0.98\textwidth]{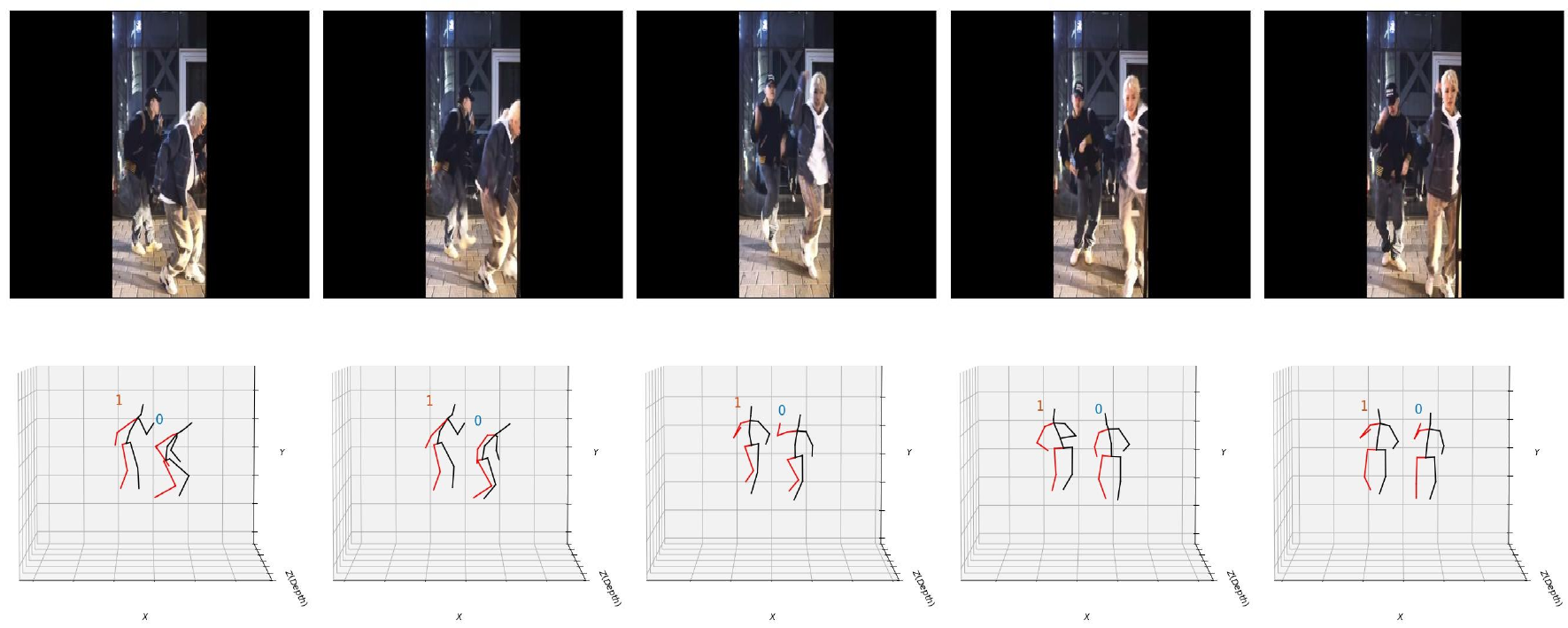}\\
  \caption{Additional Examples of in-the-wild inference (4/8) -- Dance practicing \textbf{(Padded input)}}
\end{figure*}

\clearpage

\begin{figure*} 
  \centering
  \includegraphics[width=0.98\textwidth]{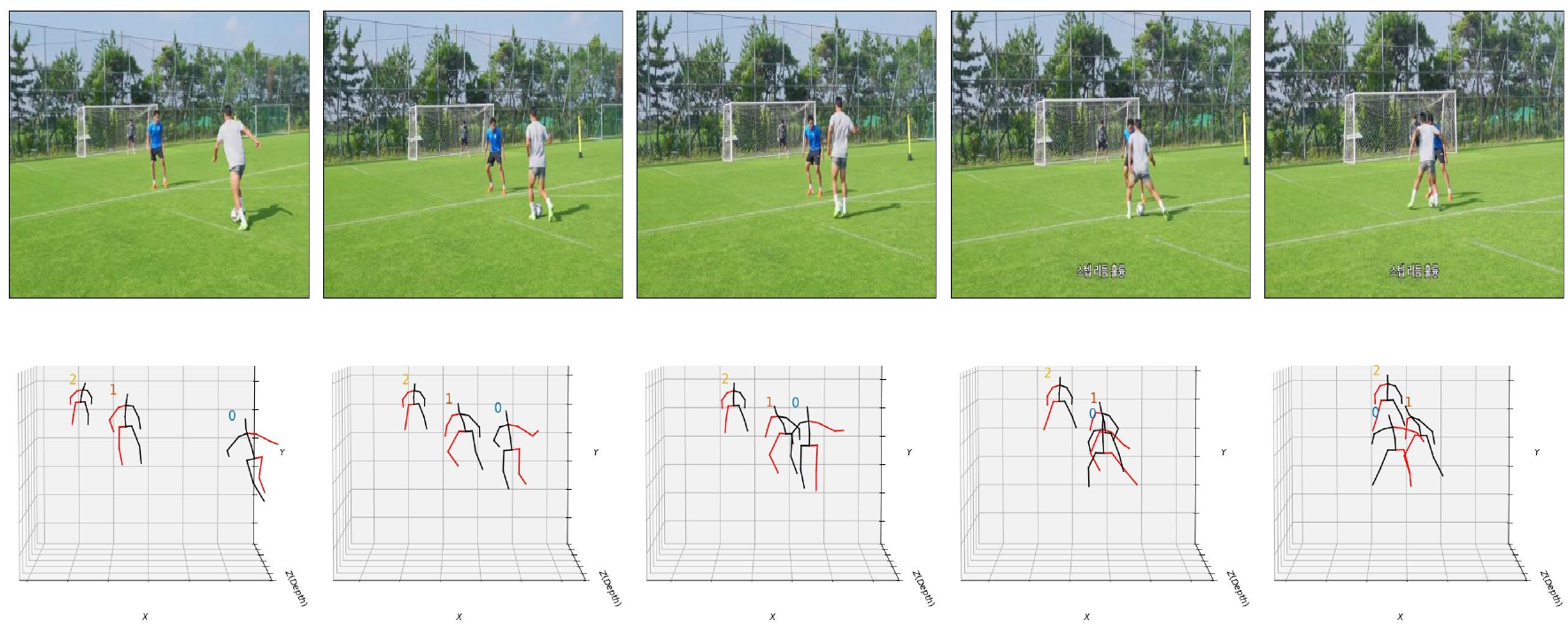}\\
  \includegraphics[width=0.98\textwidth]{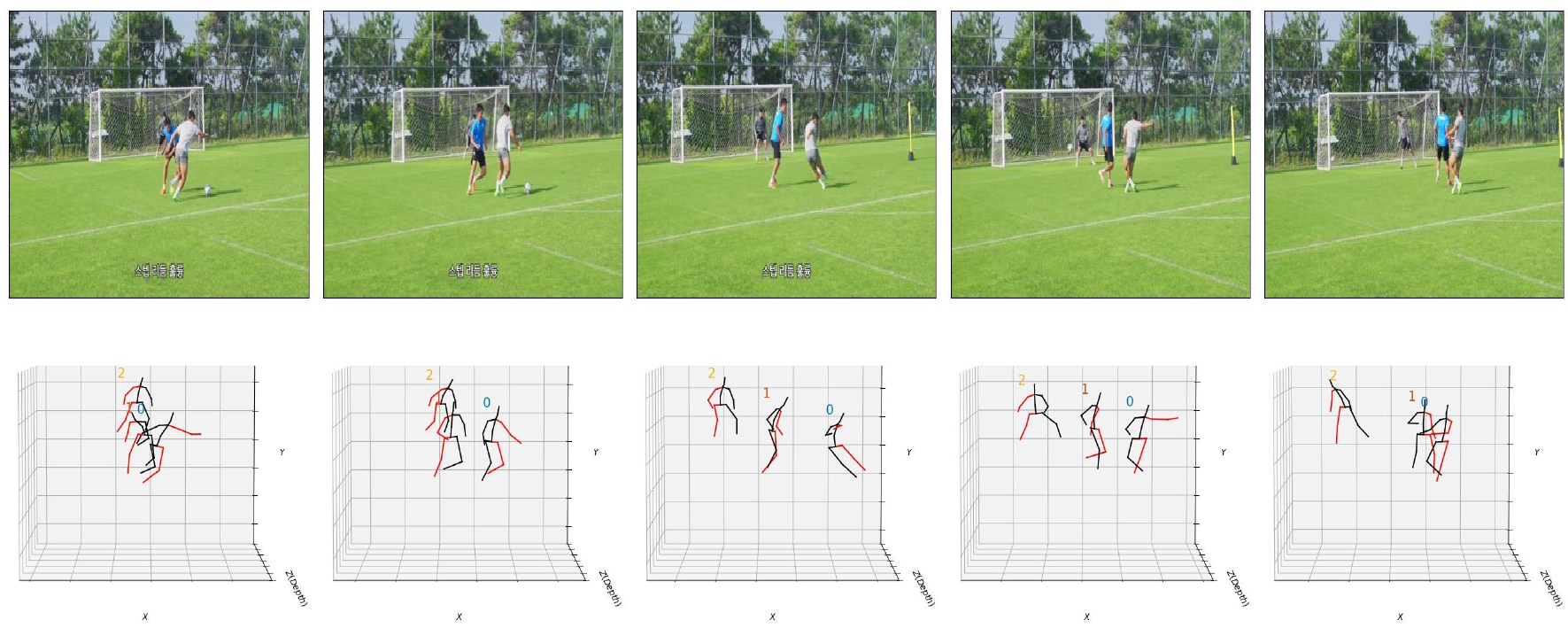}\\
  \caption{Additional Examples of in-the-wild inference (5/8) -- Professional soccer \textbf{(Rapid camera view change)}}
\end{figure*}

\clearpage

\begin{figure*} 
  \centering
  \includegraphics[width=0.98\textwidth]{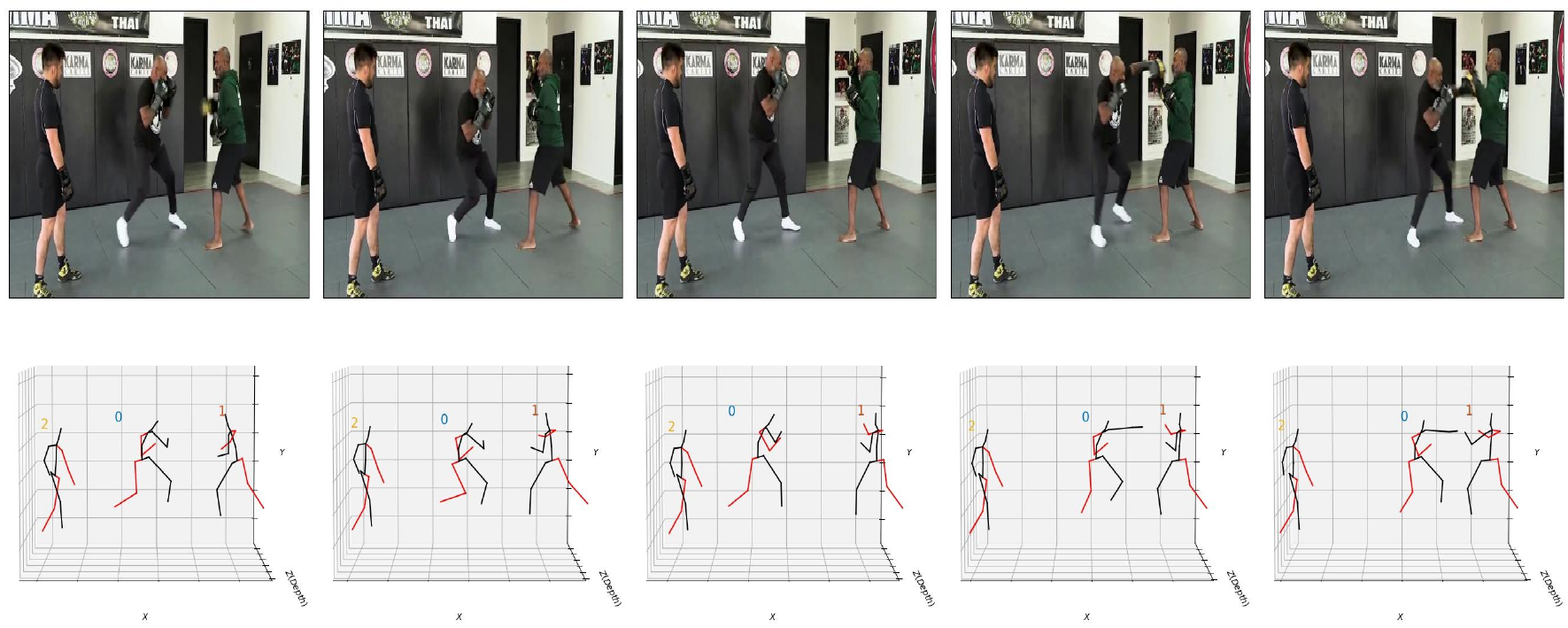}\\
  \includegraphics[width=0.98\textwidth]{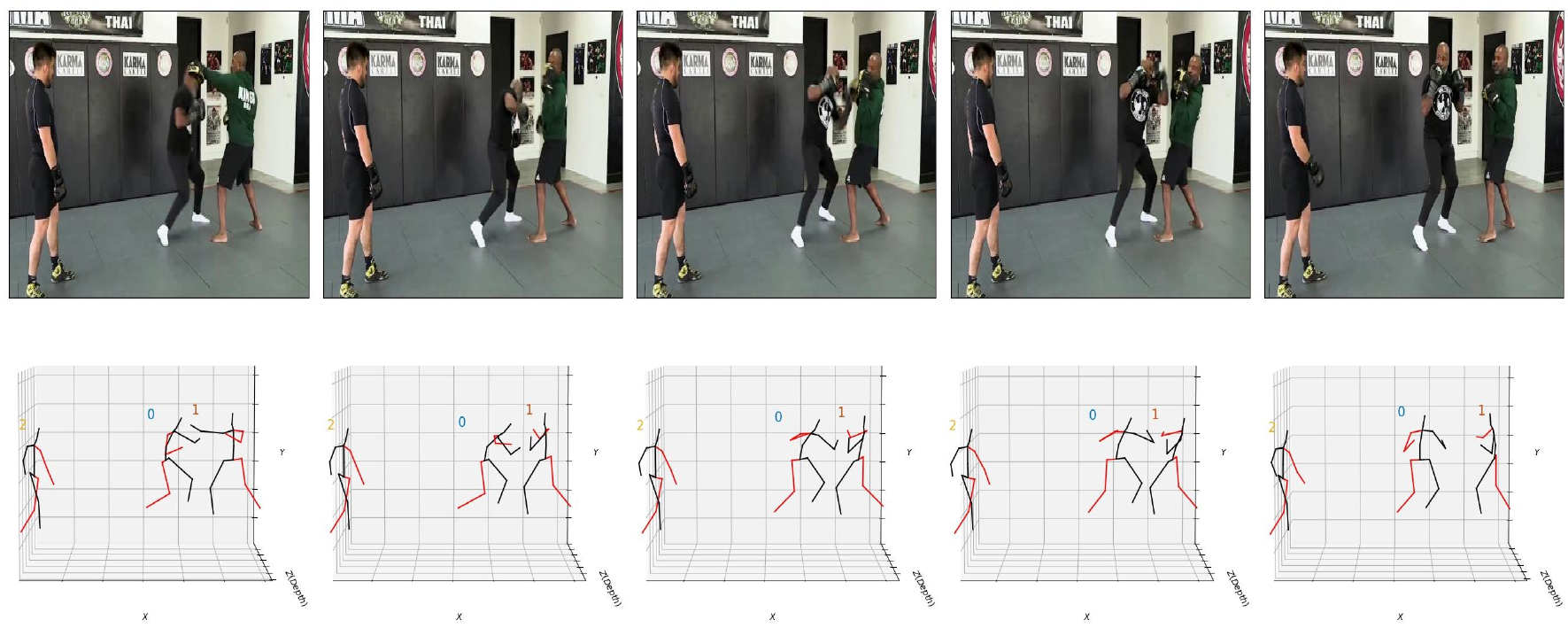}\\
  \caption{Additional Examples of in-the-wild inference (6/8) -- Professional Boxing \textbf{(Rampant movements)}}
\end{figure*}

\clearpage

\begin{figure*} 
  \centering
  \includegraphics[width=0.98\textwidth]{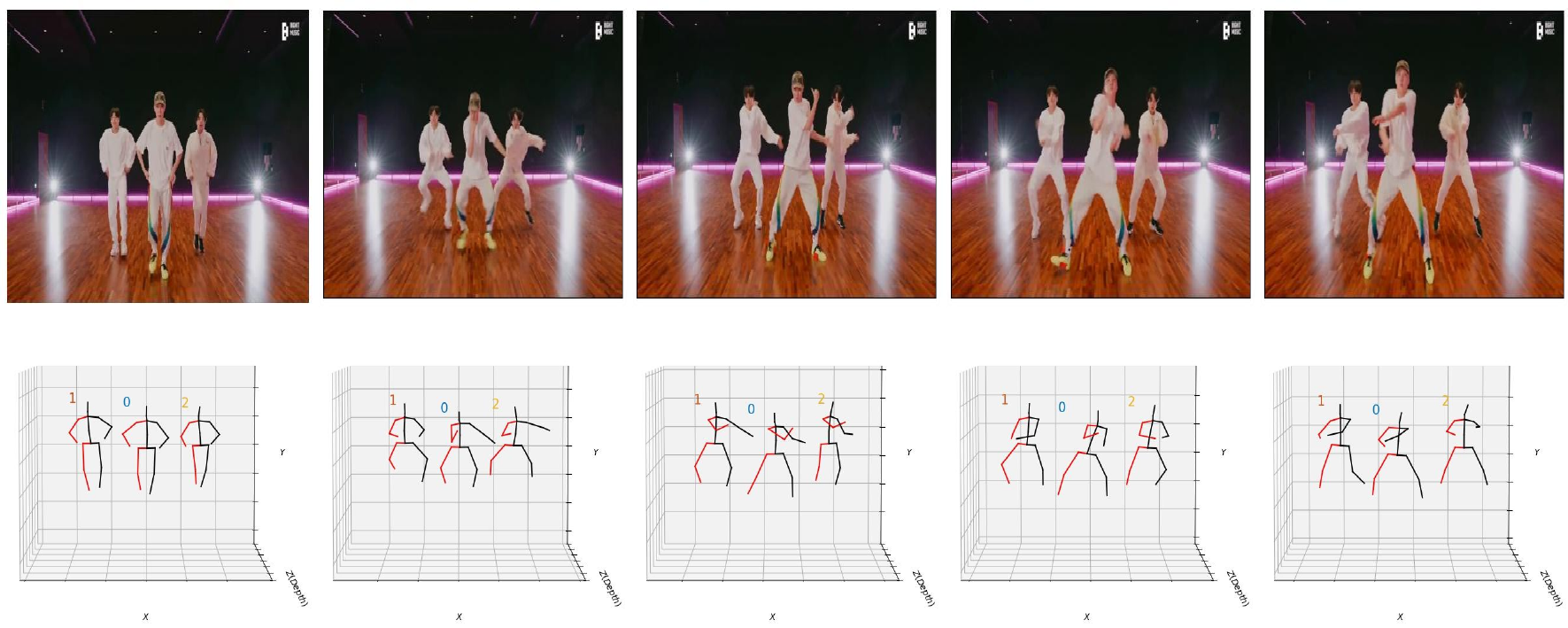}\\
  \includegraphics[width=0.98\textwidth]{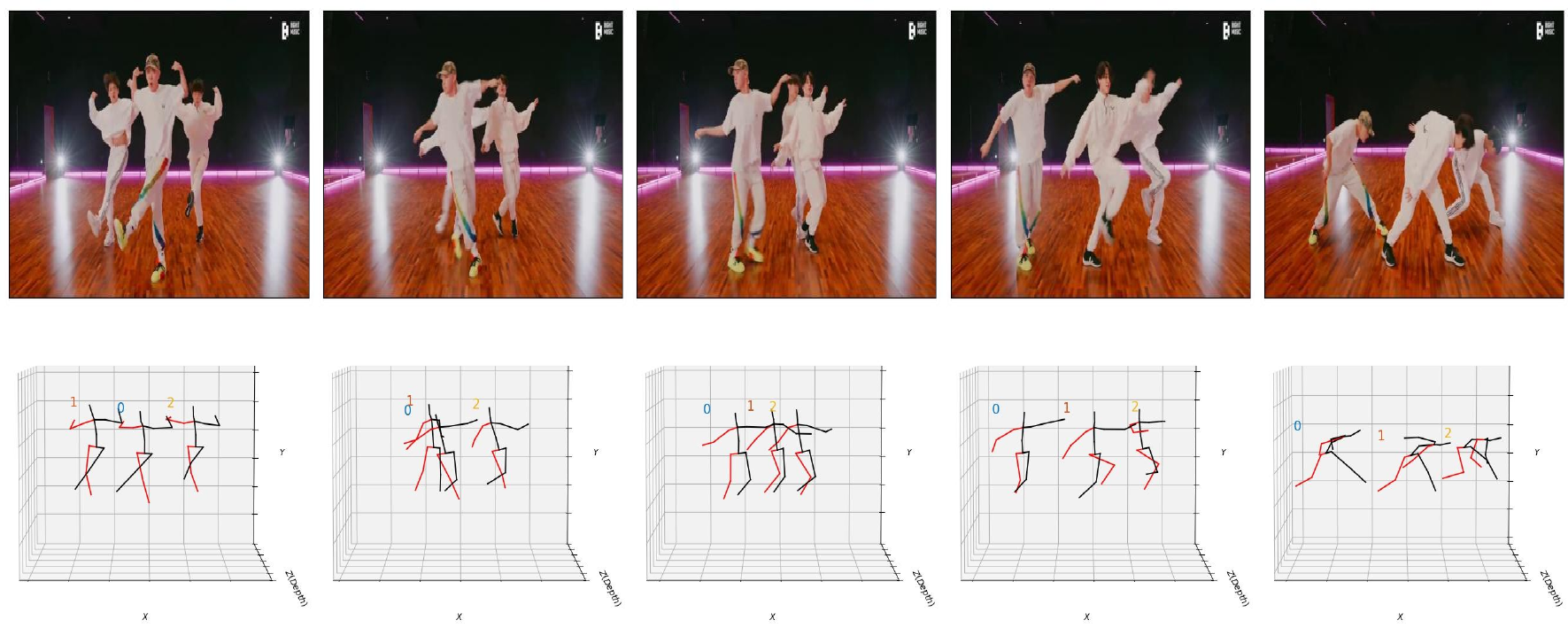}\\
  \caption{Additional Examples of in-the-wild inference (7/8) -- Boy group dance practicing \textbf{(Homogeneous Looking)}}
\end{figure*}

\clearpage

\end{document}